\documentclass[journal]{IEEEtran}
\usepackage{algorithm,url,makecell, bm}
\usepackage[caption=false,font=small,labelfont=rm,textfont=rm]{subfig}
\usepackage{amsthm}

\usepackage{xcolor,booktabs, multirow, enumitem, bm, setspace}
\usepackage{cite, amsmath, amssymb, amsfonts, algorithmic, graphicx, textcomp, etoolbox,pifont}
\allowdisplaybreaks[4]
\graphicspath{figures}

\begin{document}
\title{\huge Real-Time Communication-Aware Ride-Sharing Route Planning for Urban Air Mobility: A Multi-Source Hybrid Attention Reinforcement Learning Approach}
\author{Yuejiao Xie, Maonan Wang, Di Zhou, Man-On Pun, and Zhu Han
	\thanks{This work was supported in part by the Hong Kong Research Grants Council through the General Research Fund under Grant 17617024, and in part by the Guangdong Provincial Key Laboratory of Future Networks of Intelligence under Grant 2022B1212010001. {\em Corresponding author: Man-On Pun (SimonPun@cuhk.edu.cn)}}
	\thanks{Y. Xie, M. Wang and M.O. Pun are with School of Science and Engineering, The Chinese University of Hong Kong, Shenzhen, 518172, China.}
	\thanks{D. Zhou is with the Xidian University State Key Laboratory of Integrated Services Network, Xidian University, Xi'an, Shanxi, China.}
	\thanks{Z. Han is with the Department of Electrical and Computer Engineering at the University of Houston, Houston, TX 77004 USA, and also with the Department of Computer Science and Engineering, Kyung Hee University, Seoul, South Korea, 446-701}
}
\maketitle

\begin{abstract} 
Urban Air Mobility (UAM) systems are rapidly emerging as promising solutions to alleviate urban congestion, with path planning becoming a key focus area. Unlike ground transportation, UAM trajectory planning has to prioritize communication quality for accurate location tracking in constantly changing environments to ensure safety. Meanwhile, a UAM system, serving as an air taxi, requires adaptive planning to respond to real-time passenger requests, especially in ride-sharing scenarios where passenger demands are unpredictable and dynamic. However, conventional trajectory planning strategies based on predefined routes lack the flexibility to meet varied passenger ride demands. To address these challenges, this work first proposes constructing a radio map to evaluate the communication quality of urban airspace. Building on this, we introduce a novel Multi-Source Hybrid Attention Reinforcement Learning (MSHA-RL) framework for the challenge of effectively focusing on passengers and UAM locations, which arises from the significant dimensional disparity between the representations. This model first generates the alignment among diverse data sources with large gap dimensions before employing hybrid attention to balance global and local insights, thereby facilitating responsive, real-time path planning. Extensive experimental results demonstrate that the approach enables communication-compliant trajectory planning, reducing travel time and enhancing operational efficiency while prioritizing passenger safety.
% Our code is publicly available at https://github.com/Ricca-xie/AccessScheduling\_DRL.
\end{abstract}

\begin{IEEEkeywords}
UAM ride-sharing trajectory, Radio map, Communication connectivity, MSHA-RL
\end{IEEEkeywords}

\section{Introduction}
% 1 城市发展，交通是重要的部分
% 2 地面交通存在问题
% 3 UAM是解决交通问题的革命性方法， 
% 4 UAM的相关研究
As cities continue to expand, there is a rising demand for efficient and optimized services for transportation solutions in modern society. Ground transportation in dense urban areas has long been hindered by challenges such as congestion and limited resilience. In response, both industry and academia have started to develop low-altitude airspaces and explore new
modalities of transportation. Among these, Urban Air Mobility (UAM) stands out as a promising solution with the potential to transform urban transportation\cite{thipphavong2018urban}. High-profile initiatives in UAM, such as Joby Aviation's Electric Vertical Takeoff and Landing (eVTOL) aircraft, Amazon Prime Air, NASA's Unmanned Traffic Management program, Samson Sky's Blade, Xiaopeng's Traveler, and EHang's Autonomous Aerial Vehicle\cite{neto2021trajectory,cohen2021urban}, demonstrate the technology’s capability to redefine city travel. 
Therefore, integrating UAM with air taxis not only creates a more cost-effective and flexible alternative to traditional helicopter services, but also establishes an innovative transportation system that will revolutionize the low-altitude economy sector, paving the way for efficient and sustainable urban mobility solutions~\cite{huang2024low}.

% 拼车问题的发展
% 拼车优点：节省公里数，缓解交通拥堵，市场服务率高，降低总的时间成本
% 地面交通的研究不能直接应用，因为要考虑通信条件
% 解释空中信号不好的原因
In particular, optimizing path planning is a crucial challenge in advancing UAM technology. To maximize the carrying capacity of the UAM system for optimal energy efficiency, this work considers the ride-sharing scenario. 
Ride-sharing can significantly reduce total travel distance by allowing multiple passengers with similar routes to share a single vehicle. 
% A higher market service rate is achieved because transportation resources are utilized more efficiently, maximizing the occupancy of each vehicle.  
While ride-sharing strategies are well-established in ground transportation \cite{wang2019ridesourcing, narayanan2020shared, chen2020dynamic}, they cannot be directly applied to UAM path planning due to the challenge of maintaining reliable operations for autonomous UAM flying at high altitudes. The UAM connectivity can provide the Communication and Control (C2) for the transmissions of flight data while also supporting Payload communication (non-C2) for non-priority data transmissions \cite{piccioni2024enhancing}. A seamless and reliable UAM system can mitigate collision issues through a collision alert system. Moreover, real-time location tracking and environmental updates can be facilitated by maintaining consistent and reliable communication with the Ground Base Station (GBS), as demonstrated in \cite{cai2023safety}. In the event of an emergency, this communication link enables immediate control and response, ensuring prompt action to prevent catastrophic accidents. 

However, signal coverage at high altitudes is often less comprehensive than at ground level. GBSs are equipped with flat panel antennas that emit signals in a directional pattern, with most antennas oriented downwards to serve ground-based users \cite{challita2019machine}. In \cite{jiang2023physics}, they investigated that the channel model between UAM and GBS typically exhibits a high line-of-sight (LOS) probability in the absence of significant building occlusion. In contrast, complex urban environments often introduce non-line-of-sight (NLOS) conditions and other channel impairments, such as multipath and shadowing. As a result, airborne devices often experience reduced signal strength and coverage compared to ground-based devices, which significantly impacts the reliability and performance of communication. To address these challenges, radio maps are increasingly used to accurately characterize these wireless communication channels, providing crucial location-specific details such as channel power gain, interference, shadowing, and angles of arrival and departure as referred to \cite{mo2019radio}. Consequently, integrating a radio map is highly beneficial for UAM systems facing communication constraints.

% 1 利用radio map 优化通信约束下的UAV飞行轨迹
% 2 UAM不仅需要导航，还需要解决用户动态需求
% 3 有效的路径规划是必须的
Extensive research has been conducted in \cite{bulut2018trajectory, khamidehi2019power, zhang2020radio} on the development of optimal Unmanned Aerial Vehicle (UAV) trajectory planning strategies with communication constraints from different aspects. 
Meanwhile, a deep reinforcement learning (DRL) framework was designed for UAVs to ensure coverage-aware navigation in \cite{zeng2021simultaneous}. However, for UAM ride-sharing, these conventional communication-aware trajectory planning methods are ineffective. Specifically, UAM path planning needs to incorporate dynamic path adjustments based on real-time passenger demands, which is not addressed in traditional UAV applications.

To cope with the aforementioned challenges, this work investigates the path planning of ride-sharing for the UAM system, which completes the pick-up task of passengers and minimizes flight time while ensuring reliable communication. In particular, a radio map is constructed to simulate communication quality, and a novel multi-source hybrid attention reinforcement learning (MSHA-RL) architecture is designed, which can fuse multi-source information to generate the optimal UAM trajectory in a real-time manner. 
The main contributions of this work are summarized as follows:
\begin{itemize}[leftmargin=*]
\item The ride-sharing trajectory planning for UAM to pick up passengers in real-time under communication connectivity is first formulated. We construct the radio map to evaluate the quality of communication between UAM and GBSs;
\item A novel online architecture MSHA-RL is proposed to manage the attention of the model on passengers, UAM, and maps since there exist large dimension gaps among them. This architecture integrates multi-source features and a hybrid attention fusion module, effectively generating the alignment of multiple attributes while ensuring that more attention is given to the passenger and UAM-located area;
\item Extensive experiments are performed on two practical scenarios, and the results demonstrate that the proposed MSHA-RL architecture significantly outperforms the conventional methods with an optimized real-time trajectory.
\end{itemize}

The remainder of this paper is organized as follows. Sec.~\ref{sec:related} first presents some related works before the problem is formulated in Sec.~\ref{sec:problem}. After that, the proposed MSHA-RL architecture is provided in Sec.~\ref{sec:alg} while Sec.~\ref{sec:experiment} reports the simulation experiments conducted. Finally, conclusion is given in Sec.~\ref{sec:conclude}.

\section{Related Work}\label{sec:related}

% In this section, we provide an overview of the development of UAM and its applications, with a focus on the path planning problem.

% UAM的发展
\subsection{Development of UAM Service}
The development of UAM has been a remarkable journey driven by technological advancements and innovative concepts, such as materials engineering, battery technology, novel aircraft design, and control theory \cite{rajendran2020air}. Since the 21st century, UAM has evolved from an idea to a rapidly growing industry, with the introduction of eVTOL aircraft by NASA in 2003 \cite{moore200321st}. The successful test flight of Terrafugia's Transition in 2009 marked a significant milestone, and the introduction of the professional concept of UAM by NASA in 2017 further solidified the industry's focus on low-altitude airspace transportation\cite{straubinger2020overview}. 

A growing number of global companies are entering the UAM market, with plans to initiate commercial operations between 2024 and 2026. According to these announcements, there are two UAM service types: air-taxi and airport shuttle services \cite{coppola2024urban}. Airport shuttle services operate on fixed schedules and routes, offering a cost-effective solution, whereas air-taxi services provide customized point-to-point transportation to accommodate individual and flexible demands. In \cite{rajendran2020air}, the authors highlighted the importance of air taxi services and anticipated potential challenges, including ride-matching, pricing strategies, and real-time vehicle routing. Typically, air taxi services offer ride-sharing options, allowing multiple passengers to share a flight and split the cost, which is an attractive proposition for the UAM system. By distributing operating costs and emissions across multiple passengers, ride-sharing can make UAM more affordable and environmentally sustainable \cite{biswas2024passenger}. Therefore, 
ride-sharing route planning for air taxi services has emerged as a prominent research area, and we proceed to review the relevant literature.

%
% 路径规划问题的发展
%[12], [13] ,[14]都研究了在通信保持的情况下，优化路线设计
%[14] 还探究了该问题的复杂度。
\subsection{Trajectory Planning Techniques in UAV and UAM}
As a manned flying vehicle, UAM inherits many technical aspects of path planning from UAVs, while also exhibiting distinct differences. In UAV path planning, extensive research has been conducted on trajectory designs with consideration of disconnection duration constraints \cite{bulut2018trajectory, zhang2020radio}, emphasizing the critical importance of maintaining continuous connectivity. In addition to connectivity concerns, \cite{khamidehi2019power} investigated the complexities associated with trajectory planning for drones limited by connectivity requirements, emphasizing strategies to mitigate these challenges effectively. However, the communication model they used is relatively simple and differs significantly from real-world scenarios. In \cite{mo2019radio, zhang2019radio, dong2022radio}, radio maps were utilized for path planning under more realistic assumptions. Moreover, a comprehensive 3D path planning assisted by radio maps was illustrated in \cite{zhang2020radio}, minimizing the travel distance of UAVs while satisfying different communication requirements. The above studies all utilize traditional algorithms, such as $A^\ast$, but lack flexibility. \cite{zeng2021simultaneous} introduced a DRL framework designed specifically for cellular-connected UAVs to ensure flexible coverage-aware navigation.

\begin{figure}[htp]
	\begin{center}
	\includegraphics[width=0.35\textwidth]{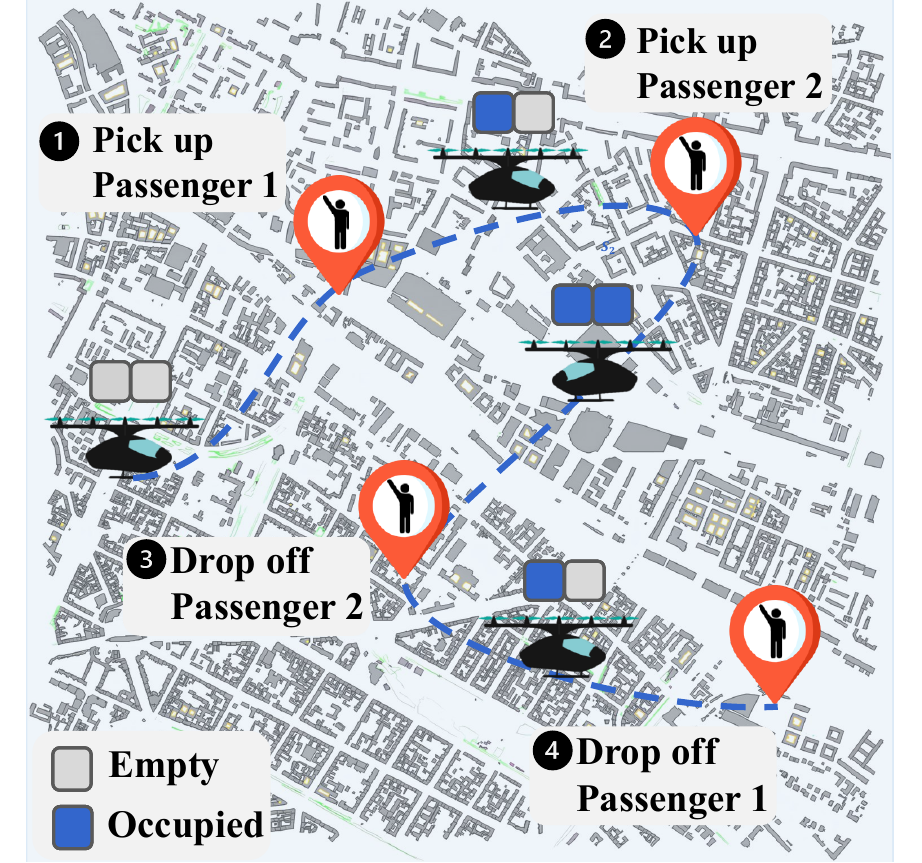}
	\end{center}
	\caption{Illustration of the UAM ride-sharing trajectory planning system.}
	\label{brackground}
\end{figure}
Unlike UAV applications, which primarily focus on navigation and surveillance, UAM provides transportation services for people or goods, typically operating at higher altitudes and faster speeds. Pang et al.\cite{pang2025v2x} focus on optimizing passenger decisions in selecting vertiports for UAM to enhance overall travel efficiency instead of flight planning. In order to mitigate the risk of collisions, \cite{tang2021automated} proposed an automated flight planning system that computes collision-free trajectories through a mixed-integer second-order cone programming approach. \cite{lou2021rrt} explored a sampling-based trajectory planning approach using rapidly exploring random trees (RRT*) to find the cost-optimal path. However, these methods are based on fixed environments, which limit their robustness in real-world scenarios. ~\cite {yang2021autonomous} investigated collision-free trajectory planning with uncertain obstacles by using the Monte-Carlo tree search method. Nevertheless, this framework did not consider the need for communication connectivity. While Xie et al.~\cite{xie2025dynamic} explored a communication-aware UAM path planning method, their study is limited to a small urban scenario.
These architectures have good extensibility and collectively advance our understanding of optimal UAM trajectory design across multiple dimensions.

\section{System Model and Formulation}\label{sec:problem}
In this section, we will first introduce the UAM system, a novel transportation concept envisioned as an on-demand air taxi service, designed to provide efficient and sustainable transportation solutions for urban areas. After that, we will formulate the corresponding path-planning problem, which aims to optimize the routing and scheduling of UAM vehicles to minimize travel times, reduce congestion, and enhance overall system efficiency.

\begin{table}[htp]
	\centering
	\caption{Definition of Notation}
	
	\begin{tabular}{ll} 
		\toprule
		\makecell[l]{\textbf{Notations}} & \makecell[c]{\textbf{Definition}} \\ \hline
		$\mathcal{S}(t)$ & The pick-up point set at time $t$. \\
		$\mathcal{D}(t)$& The drop-off point set at time $t$.\\
		$I(t)$ & The passenger indice set at time $t$. \\
		$N$ & The total number of passengers. \\
		$T$ & The mission completion time. \\
		$\boldsymbol {u}(t)$ & \makecell[l]{The coordinate of location at time $t$.} \\
		$\boldsymbol {u}_A(0)$ & The starting point of UAM at initial time.\\
		$\boldsymbol {u}_q(t)$ &\makecell[l]{The pick-up or drop-off point in \\$\{\mathcal{S}(t), \mathcal{D}(t)\}$ at time $t$.}\\
		$\boldsymbol {u}_{S_i}(t)$ &The pick-up point in $\mathcal{S}(t)$ at time $t$.\\
		$\boldsymbol {u}_{D_i}(t)$ &The drop-off point in $\mathcal{D}(t)$ at time $t$.\\
		$\tau$ & The point in time.\\
		$V$ & The speed of UAM.\\
		$U_{\text{seats}}(t)$ & \makecell[l]{The number of remaining available seats \\at time $t$.}\\
		$x(t)$& The x coordinate at time $t$. \\
		$y(t)$& The y coordinate at time $t$. \\
		$L$ & The boundary of UAM flight.\\
		$H$    & The altitude of UAM flight.\\
		$\bar{\gamma}$ &The expected SINR.\\
		$\boldsymbol{U}(t)$ & \makecell[l]{A series of UAM trajectories until time $t$.}\\
		$\boldsymbol{U}_{\text{seats}}(t)$ & \makecell[l]{A series of the number of remaining \\ available seats until time $t$.}\\
		$S_n(t)$ & \makecell[l]{the starting point of passenger $n$, $n \in I(t)$.}\\
		$D_n(t)$ & \makecell[l]{the destination point of passenger $n$, \\$n \in I(t)$.}\\
		$d_n(t)$ & \makecell[l]{the distance of the $n$-th passenger from UAM, \\$n \in I(t)$.}\\
		$\alpha_n(t)$ & \makecell[l]{The status of whether the $n$-th passenger is \\ on board, $n \in I(t)$.}\\
		$\beta_n(t)$ & \makecell[l]{The status of whether the $n$-th passenger is served, \\$n \in I(t)$.}\\
		$\bar{\Gamma}(t)$ & \makecell[l]{A range of SINR map at time $t$.}\\
		$\bar{E}(t)$ & \makecell[l]{A range of uncertainty map at time $t$.}\\
		\bottomrule
	\end{tabular}
	
	\label{tab:my_label}
	% \vskip -1em
\end{table}
\subsection{UAM Service Scenario}
Considering a cellular-connected UAM as a complement to ground transportation, our goal is to optimize its trajectory to satisfy the real-time requests of passengers while prioritizing communication during flight, as illustrated in Fig.~\ref{brackground}. 
%定义坐标和路线，以及速度
The mission of UAM is to transport passengers from their starting point to their destination. In the Cartesian coordinate system, the coordinate of a location at time $t$ is defined as $\boldsymbol {u}(t) = \left [x(t), y(t), H(t)\right ]$. Furthermore, the initial starting point of UAM can be denoted as $\boldsymbol {u}_{A}(0) = \left [x(0), y(0), H(0)\right ]$. As passengers arrive dynamically over time, the pick-up points are described by the time-dependent set $\mathcal{S}(t) = \{ S_i | i \in I(t)\}$, while destinations are also represented as $\mathcal{D}(t) = \{D_i | i \in I(t) \}$, where $|{I}(t)|\leq N, t \leq T$. Here, $I(t)$ is the set of passenger indices that require service at time $t$, with the total number of passengers $|{I}(t)|$ limited to $N$, and $T$ denotes the completion time of the mission. Upon successful transportation of all passengers to their designated destinations, the UAM achieves its objectives. All the take-off and landing points mentioned are airports where UAM can be parked or recharged. We assume there is more than one GBS, i.e., $M>1$, with the $m$-th GBS located at coordinate $\boldsymbol {u}_{m}$. The notations used throughout this paper are summarized in Table~\ref{tab:my_label}, along with their corresponding definitions.

\subsection{Problem Formulation}
The objective of this study is to minimize the total travel time $T$ while ensuring safety and successfully transporting all passengers to their destinations. Based on the area radio map, the UAM path should avoid regions with low signal strength to ensure reliable communication. Therefore, the task of reducing the completion time $T$ by optimizing the UAM path $\boldsymbol{u}(t)$ can be formulated as follows:

\begin{align}
&~~\mathrm{P} : \min_{T, \boldsymbol{u}(t)} T, \\
\text {s.t.} &~~\bar{\gamma}(\boldsymbol{u}(t)) \geq \bar{\gamma}_{\mathrm{T}},  ~ \forall t \in[0, T], \label{constraint:R} \\
&~~ \boldsymbol{u}(0) = \boldsymbol{u}_A(0), \label{constraint:start}\\
&~~ \forall \boldsymbol {u}(T) \in \mathcal{D}(T),\label{constraint:end} \\
&~~ \boldsymbol{u}(t)=\boldsymbol {u}_q(t),  ~\forall q \in\left\{\mathcal{S}(t), \mathcal{D}(t)\right\}, \exists t \in[0, T],\label{constraint:passen} \\
&~~\tau \left(\boldsymbol {u}_{S_i}\right)<\tau \left(\boldsymbol {u}_{D_i}\right), ~ \exists t \in[0, T], \forall i \in I(t), \label{constraint:time}\\
&~~V(\boldsymbol{u}(t))=\bar{V}, \forall t \in[0, T], \label{constraint:speed}\\
&~~ 0 \leq U_{\text{seats}}(t) \leq N_{\text {seat}}, \forall t \in[0, T], \label{constraint:seats}\\
&~~0 \leq x(t) \leq L, \forall t \in[0, T], \label{constraint:x} \\
&~~0 \leq y(t) \leq L, \forall t \in[0, T], \label{constraint:y} \\
&~~H(\boldsymbol{u}(t))=\bar{H}, \forall t \in[0, T], \label{constraint:h}
\end{align}
where Eq.~\eqref{constraint:R} sets a signal-to-interference-plus-noise ratio (SINR) threshold $\bar{\gamma}_{\mathrm{T}}$ along the UAM's task execution path, maintaining the expected SINR $\bar{\gamma}(\boldsymbol{u}(t))$ at location $\boldsymbol{u}(t)$ to stay above the minimum threshold to ensure communication requirements. Eqs.~ \eqref{constraint:start} and \eqref{constraint:end} define the initial and final points of the UAM, respectively. After completing the transport mission, the UAM can park at the airport where the last passenger is located. Eq.~\eqref{constraint:passen} indicates that all passengers must be picked up and transported to the destination to fulfill the task. $q$ represents the pick-up or drop-off point of passenger at time $t$, and $\boldsymbol {u}_q(t)$ denotes its coordinate. Eq.~\eqref{constraint:time} restricts the UAM to arrive at the passenger's starting point before reaching the corresponding destination, and $\tau$ represents the point in time. Eq.~\eqref{constraint:speed} denotes that the UAM keeps a constant speed $\bar{V}$ during the flight. Eq.~\eqref{constraint:seats} shows that the number of remaining available seats indicated by $U_{\text{seats}}(t)$ cannot exceed the available seating capacity $N_{\text {seat}}$. Finally, Eqs.~\eqref{constraint:x} and \eqref{constraint:y} define the boundaries of the UAM flight, while Eq.~\eqref{constraint:h} indicates that it flies at a constant altitude $\bar{H}$.

\section{Radio Map Construction}\label{sec:map}
To evaluate the communication quality of the airspace, we propose to adopt the radio map. We will first define the communication model between UAM and GBS, followed by demonstrating the construction process of the SINR map.

\begin{figure*}[htbp]
	\begin{center}
		\includegraphics[width=0.8\textwidth]{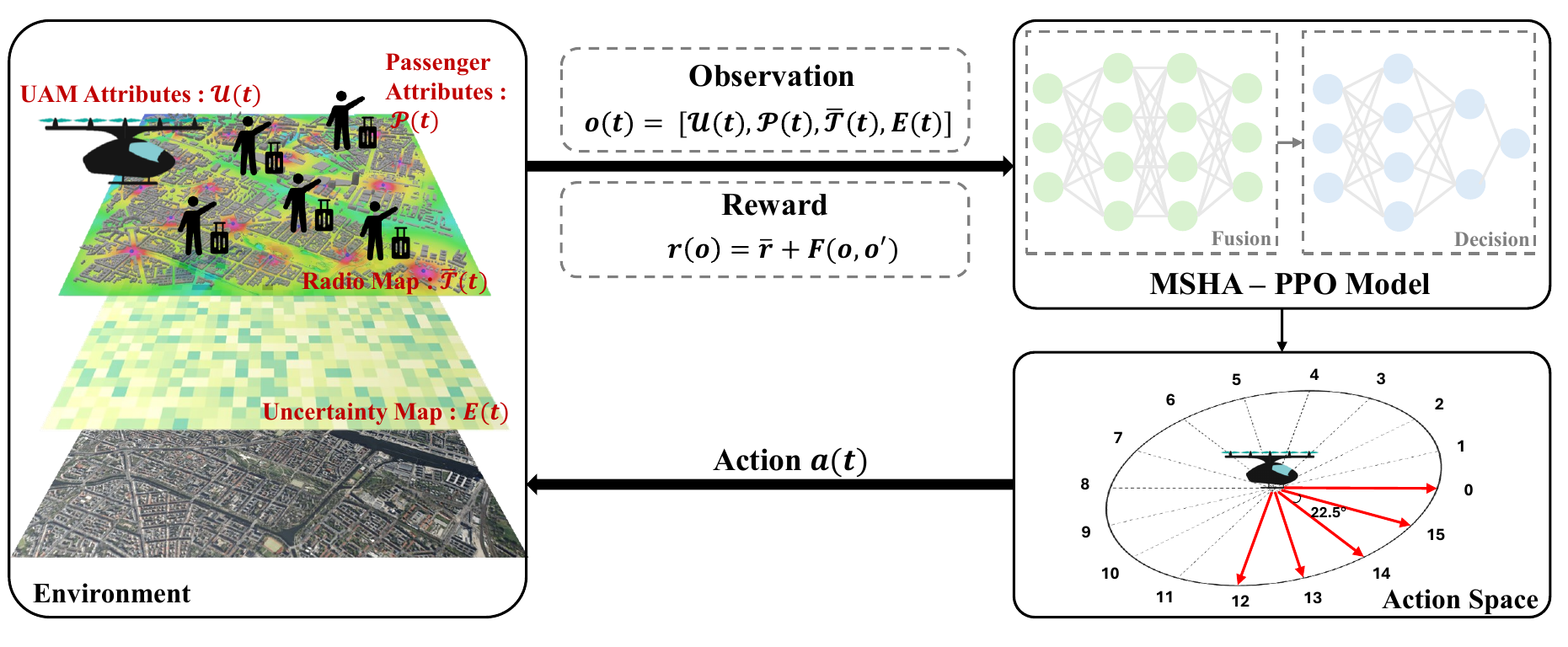}
	\end{center}
	\caption{The overall architecture of Multi-Source Hybrid Attention-Reinforcement Learning.}
	\label{RL_architecture}
\end{figure*}

\subsection{Communication Model}
When UAM connects to GBSs, the channel gain is characterized by the large-scale channel gain and the small-scale channel fading gain, denoted as $\bar{h}_m(\boldsymbol{u})$ and $\tilde{h}_m(\boldsymbol{u})$, respectively. Therefore, the instantaneous channel gain between the $m$-th GBS and the UAM at location $\boldsymbol{u}$ can be represented by the following expression \cite{chen2024optimal}:
\begin{equation}
h_m(\boldsymbol{u})=\bar{h}_m(\boldsymbol{u}) \tilde{h}_m(\boldsymbol{u}).
\end{equation}

At any given time $t$, where $0 \leq t \leq T$, the UAM is assumed to be connected to a specific GBS at location $\boldsymbol{u} = \boldsymbol{u}(t)$. The GBS is identified by the index $G(\boldsymbol{u}) \in \mathcal{M} $, with $\mathcal{M} = \left\{1,..., M\right\}$. Considering the interference from other GBSs serving their terrestrial users, we employ the SINR to quantify the communication quality. The SINR of connection with the $G(\boldsymbol{u})$-th GBS at location $\boldsymbol{u}$ is defined as follows \cite{zhang2020radio}:
\begin{equation}
\begin{aligned}
\gamma(\boldsymbol{u}) =\frac{P h_{G(\boldsymbol{u})}^2(\boldsymbol{u})}
{P \displaystyle\sum_{m^{\prime} \in \mathcal{M} \backslash \{ G(\boldsymbol{u})\}} c_{m^{\prime}}h_{m^{\prime}}^2(\boldsymbol{u}) +\sigma^2 },
\end{aligned}
\end{equation}
where $P$ represents the transmission power of each GBS and $\sigma^2$ denotes the noise power at the UAM receiver. $c_{m^{\prime}}$ indicates the occupancy status of the frequency band by the $m^{\prime}$-th GBS, with $c_{m^{\prime}} = 1$ denoting that the $m^{\prime}$-th GBS reuses the same frequency band as the $m$-th GBS; otherwise, it is set to zero.

Considering the impact of small-scale fading and time-varying interference, the {\em expected} SINR provides a more accurate representation of communication quality than the instantaneous SINR. Furthermore, we assume that  $c_{m^{\prime}}$ follows a Bernoulli distribution, with a mean value of $l_m$, i.e., $E\left[c_{m^{\prime}}\right] = l_{m^{\prime}}$. Specifically, the estimation of $l_{m^{\prime}} \in \left[0, 1\right]$ can be determined by calculating the ratio of the average number of users served by each GBS to the total number of available frequency bands. Consequently, the expected SINR between the UAM and its corresponding GBS is given by \cite{zhang2020radio}:
\begin{equation}
\bar {\mathcal{\gamma}}(\boldsymbol{u})  
\triangleq \max_{m \in \mathcal{M}} \frac{P \bar {h}_{m}^2(\boldsymbol{u})}
{P \displaystyle\sum_{m^{\prime} \in \mathcal{M} \backslash \{ m\} } l_{m^{\prime}} \bar {h}_{m^{\prime}}^2(\boldsymbol{u}) 
+\sigma^2 }.
\label{eq:E_sinr}
\end{equation}

\subsection{SINR Map Construction}
Next, we consider a scenario where the UAM system provides passenger services within a square area at a constant altitude. To ensure computational tractability, we partition the square domain into a grid with a specified granularity $\Delta_C$, with the number of discretized cells denoted by $ N_C \times N_C $, where $N_C=L/\Delta_C$. By selecting $\Delta_C$ infinitesimally small, we can assume that the SINR values remain approximately constant. 
The grid cell $\boldsymbol{u}_C(i,j)$ is can be defined as:
\begin{equation}
\begin{aligned}
    \boldsymbol{u}_C\left(i, j\right)=\boldsymbol{u}_R+\left[i-1, j-1\right]^T \Delta_C,
\end{aligned}
\end{equation}
where $i,j \in \left\{1,2,\dots, \mathcal{L}\right\}$ and $\boldsymbol{u}_R \in \mathbb{R}^{2 \times 1}$ is the minimum coordinate in the 2D space. 

Therefore, we can determine the expected SINR according to Eq.~\eqref{eq:E_sinr}:
\begin{equation}
\bar {\gamma}(\boldsymbol{u}_C (i,j))  
= \max_{m \in \mathcal{M}} \frac{P \bar {h}_{m}^2\left(\boldsymbol{u}_C (i,j)\right)}
{\tilde{\sigma}^2\left(\boldsymbol{u}_C (i,j)\right)- P l_m \bar {h}_{m}^2\left(\boldsymbol{u}_C (i,j)\right)}, 
\end{equation}
where 
\begin{equation}
\tilde{\sigma}^2\left(\boldsymbol{u}_C (i,j)\right) \triangleq \sigma^2 + P\sum_{m^{\prime} \in \mathcal{M}} l_{m^{\prime}} \bar{h}_{m^{\prime}}^2(\boldsymbol{u}_C (i,j)).
\end{equation} 

\section{Multi-Source Hybrid Attention Fusion Enhanced Reinforcement Learning Architecture}\label{sec:alg}

In this section, we first formulate the trajectory planning problem as a Markov Decision Process (MDP), and then introduce our MSHA-RL framework, as illustrated in Fig.~\ref{RL_architecture}. Initially, we gather data from various sources, including map information, UAM attributes, and passenger attributes. The collected information is then processed by the MSHA-RL framework, which consists of two modules, namely a feature extraction and fusion module that integrates multi-source features, and a decision module that determines action based on the output from the feature extractor. After the execution of the action, a reward is generated, which can be used to update the overall policy.

\subsection{MDP Framework}
Since the travel demand of passengers changes over time, we model the problem as an MDP, where the self-navigating agent explores the optimal trajectory to complete the task of picking up passengers. The MDP is characterized by the tuple $(O, A, R, \mathrm{Tr})$ where $O$ denotes the state vector, which integrates information from multiple sources, including UAM attributes, passenger attributes, radio maps, and uncertainty maps. Furthermore, $A$ represents selectable flight directions for UAM at each step, whereas $R:O \times A \rightarrow \mathbb{R}$ is the reward function that measures the immediate utility of actions. Finally, $\mathrm{Tr}:O \times A \times O \rightarrow [0,1]$ is the transition function, which characterizes the transition probability between states based on the actions executed.

As shown in Fig.\ref{RL_architecture}, during each time step $t$, UAM collects multi-source observations and inputs to the MSHA-RL framework to extract features. The overall observation at time $t$ denoted by $o(t)$ includes UAM attributes $\mathcal{U}(t)$, passenger attributes $\mathcal{P}(t)$, radio maps $\bar{\Gamma}(t)$, and uncertainty map $E(t)$ and is defined as:
\begin{equation}
    o(t) = [\mathcal{U}(t),\mathcal{P}(t),\bar{\Gamma}(t),E(t)].
\end{equation} 

The UAM's attributes defined as $\mathcal{U}(t) = [U(t), U_{\text{seats}}(t)]$ consist of the historical motion trajectories of the aircraft within the last several time steps, i.e., $U(t) \in \mathbb{R}^{t \times 2}$, and the number of available seats denoted as $U_{\text{seats}}(t) \in \mathbb{R}^{t \times 1}$. Also, the UAM system receives different ride requests at different time slots. For each passenger $n \in I(t)$, the key 
attributes $p_n(t) \in \mathcal{P}(t)$ can be expressed as: 
\begin{equation}
    p_n(t)=[S_n(t), D_n(t), d_n(t), \alpha_n(t),\beta_n(t)],
\end{equation}
where $S_n(t), D_n(t) \in \mathbb{R}^{1 \times 2}$ are the starting point and destination point of the $n$-th passenger, $d_n(t)$ represent the Euclidean distance of the $n$-th passenger from the UAM, and $\alpha_n(t),\beta_n(t)$ respectively denote two statuses whether the the $n$-th passenger is onboard or has been served, captured by binary vectors at time $t$. To measure the quality of communication, the radio map is constructed to obtain the value of SINR. We assume that the UAM can access a range of SINR maps for the current location, expressed as $\bar{\Gamma}(t) \in \mathbb{R}^{10 \times 10}$. Finally, the uncertainty map is introduced to dynamically record the exploration degree of the UAM for a region. Specifically, it is defined as the ratio of the number of times that the aircraft has explored the region to the total number of flight steps. The partial map $E(t) \in \mathbb{R}^{10 \times 10}$ is utilized for reducing computational costs.

Through processing the multi-source information, MSHA-RL can select an action $a(t) \in \{0,1,\dots,k\}$, where $k$ represents the number of possible directions for the next UAM movement. A larger value of $k$ results in smoother and more optimal trajectories, but increases the dimension of the action space. Unless otherwise specified, we set $k=15$ throughout this work, i.e., adjacent indexes representing flight direction offsets of $22.5$ degrees as shown in Fig.~\ref{RL_architecture}. Using a constant speed, each step flies an equal distance. 
\begin{figure*}[htbp]
	\begin{center}
		\includegraphics[width=0.8\textwidth]{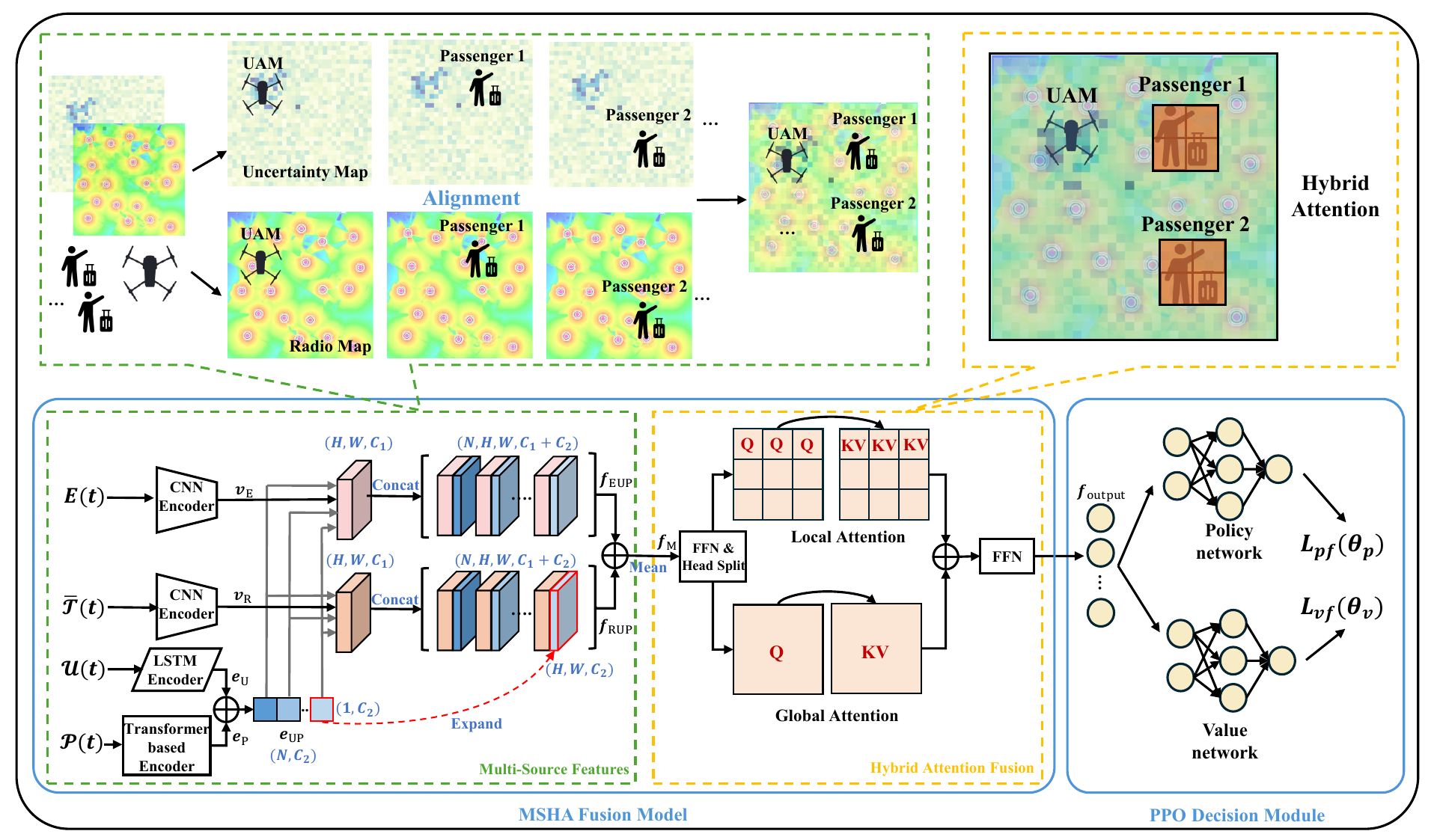}
	\end{center}
	\caption{The framework of Multi-Source Hybrid Attention-Proximal Policy Optimization Model.}
	\label{MSHA}
\vskip -1em
\end{figure*}

Following the UAM's action, the environment state transitions from $o(t)$ to $o(t+1)$ according to the transition probability $\mathrm{Tr} [o(t+1)|o(t), a(t)]$, receiving a reward $r(t) \in R$, which is then fed into MSHA-RL for model updates. To accelerate convergence in the presence of sparse reward, we propose the following shaped reward function. Specifically, we denote the reward at time $t$, $r(t)$, under state $o(t)$ as $r(o)$, and the reward at time $t+1$, $r(t+1)$, as $r(o^{\prime})$. Therefore, the shaped reward $r(o)$ is expressed as the task completion reward $\bar{r}$ plus the additional reward function $F$, as follows:
\begin{equation}
    r(o) = \bar{r} + F(o, o^{\prime}),
\end{equation}
where $F:o \times o \rightarrow \mathbb{R}$ takes the following form:
\begin{equation}
    F(o, o^{\prime}) = \zeta \cdot\Phi(o^{\prime}) - \Phi(o),
\end{equation}
with $\zeta$ denoting a discount factor. Furthermore, $\Phi:o  \rightarrow \mathbb{R}$ is a potential function, a heuristic measure of the value of each state $o$. 

The potential function is composed of incentives and penalties. By providing a pick-up reward $z_s(o)$ and a delivery reward $z_d(o)$, the function encourages the agent to serve passenger requests efficiently. Conversely,  the function penalizes the agent for undesirable outcomes, including the following three components: the flying time cost denoted by $z_{\text{time}}(o)$, the penalty for repeated exploration $z_{\text{rep}}(o)$, which is related to the number of repeated explorations, and the penalty for arriving at the region where the SINR is lower than the threshold, denoted by $z_{\gamma}(o)$. Therefore, the potential function is represented as:
\begin{equation}
    \Phi(o) = \omega_1 z_s(o) + \omega_2 z_d(o) - \omega_3 z_{\text{time}}(o)-  \omega_4 z_{\text{rep}}(o) -\omega_5 z_{\gamma}(o),\\
\end{equation}
where $\left\{\omega_i\right\}$, $i=1,2,\cdots,5$, are the weighting coefficients.

% 我们将乘客信息和飞机信息分别融入到地图信息中 
% 增加目的 To ...
\subsection{MSHA Fusion Model}
Since the states from multiple sources have different dimensions, there is a significant disparity between them, with maps having larger dimensions compared to the attributes of UAM and passengers. As a result, simply concatenating these attributes would result in those with fewer dimensions being given less importance, making it difficult to achieve proper alignment between the information. To address this issue, we propose a novel MSHA model, which involves two main modules, namely a multi-source feature module and a hybrid attention fusion. 

As depicted in the upper part of Fig.~\ref{MSHA}, the multi-source features module aligns each passenger and UAM with the corresponding radio map and uncertainty map to generate their respective feature maps. These maps are then concatenated into a comprehensive feature set whose mean is computed across channel dimensions. The subsequent hybrid attention module refines these features by utilizing hybrid attention mechanisms, allowing it to focus more on areas where passengers or UAM are located while maintaining a global perspective. The detailed workflow, as illustrated in the lower part of Fig.~\ref{MSHA}, is described in the following section.

Given multiple sources of information, we first use specific encoders to individually extract feature maps. Each user or UAM vector is then fused with the corresponding radio and uncertainty maps of the current location range. As a result, the multi-source feature extraction module produces a multi-source feature map with proper alignment. Next, the hybrid attention fusion module refines the combined feature map, capturing fine-grained details while preserving the global receptive field. Finally, the self-attentive feature maps emphasize the significance of the inputs for decision-making.

\subsubsection{Multi-Source Features}
The uncertainty map $E$ and radio map $\bar{\Gamma}$ are processed through separate convolutional neural network (CNN) \cite{lecun1998gradient} encoders to extract spatial features, $\boldsymbol{v}_\mathrm{E}$ and $\boldsymbol{v}_\mathrm{R}$, both with dimensions of $(H, W, C_1)$,
providing essential information about the environmental conditions and potential uncertainties.  The encoder process is given as:
\begin{eqnarray}
    \boldsymbol{v}_\mathrm{E} &=& \operatorname{Relu}(\operatorname{Conv}(E)),\\
    \boldsymbol{v}_\mathrm{R} &=& \operatorname{Relu}(\operatorname{Conv}(\bar{\Gamma})),
\end{eqnarray}
where $\operatorname{Relu}(\cdot)$ is an activate function.

Meanwhile, the UAM attributes $\mathcal{U}$, which are time-series data, are fed into an LSTM \cite{hochreiter1997long} encoder to capture temporal dependencies, producing $\boldsymbol{e}_U \in \mathbb{R}^{N_1 \times C_2}$ as follows:
\begin{equation}
    \boldsymbol{e}_U = \operatorname{LSTM}(\mathcal{U}),
\end{equation} 
where $N_1$ is the selected number of time steps. 

Moreover, the dynamic passenger attributes $\mathcal{P}$ are processed by a multi-head attention (MHA) encoder of transformer \cite{vaswani2017attention} to obtain comprehensive relationships of all requests. The resulting feature map $\boldsymbol{e}_\mathrm{P} \in \mathbb{R}^{N_2 \times C_2}$ can be derived as:
\begin{equation}
    \boldsymbol{e}_\mathrm{P} = \operatorname{MHA}(\mathcal{P}),
\end{equation}
with $N_2$ as the total number of passengers. 

Next, the passenger and UAM embedding are concatenated along the first dimension, yielding $\boldsymbol{e}_\mathrm{UP} \in \mathbb{R}^{N \times C_2}$, where $N=N_1+N_2$. Each vector $e \in \boldsymbol{e}_\mathrm{UP}$ is first expanded to $(H, W, C_2)$, denoted as $e^{\prime}$ and subsequently appended to the uncertainty and radio maps, yielding $\boldsymbol{f}_\mathrm{EUP}$ and $\boldsymbol{f}_\mathrm{RUP}$, both with dimension $(N, H, W, C_3)$, where $C_3 = C_1+C_2$. After that, features are aggregated before being averaged, forming $\boldsymbol{f}_\mathrm{M} \in \mathbb{R}^{H \times W \times C_3}$. Mathematically, the whole process can be written as:
\begin{equation}
    \boldsymbol{e}_\mathrm{UP} = \operatorname{Concat}(\boldsymbol{e}_\mathrm{U}, \boldsymbol{e}_\mathrm{P}),
\end{equation}
\begin{equation}
\begin{aligned}
    \boldsymbol{f}_\mathrm{EUP} = \operatorname{Concat}([{e^{\prime}}^1,\boldsymbol v_\mathrm{E}^1],\dots, [{e^{\prime}}^N,\boldsymbol v_\mathrm{E}^N]),
\end{aligned}
\end{equation}
\begin{equation}
\begin{aligned}
    \boldsymbol{f}_\mathrm{RUP} = \operatorname{Concat}([{e^{\prime}}^1,\boldsymbol v_\mathrm{R}^1],\dots, [{e^{\prime}}^N,\boldsymbol v_\mathrm{R}^N]),\\
\end{aligned}
\end{equation}
\begin{equation}  
\boldsymbol{f}_\mathrm{M} = \operatorname{Mean}(\operatorname{Concat}(\boldsymbol{f}_\mathrm{EUP}, \boldsymbol{f}_\mathrm{RUP})),
\end{equation}
where $\operatorname{Concat}(\cdot)$ and $\operatorname{Mean}(\cdot)$ denote the concatenate and mean operations, respectively.

The alignment of UAM vehicles and passengers with maps has been integrated as $\boldsymbol{f}_\mathrm{M}$.

\subsubsection{Hybrid Attention Fusion}
The designed hybrid attention fusion then follows the multi-source feature module. The feature $\boldsymbol{f}_\mathrm{M}$ is first processed by a Feed-Forward Network (FFN) layer, forming $\boldsymbol{f}_\mathrm{M}' \in \mathbb{R}^{H_{\text{head}} \times HW \times \frac{C_3}{H_{\text{head}}}}$:
\begin{equation}
    \boldsymbol{f}_\mathrm{M}' = \operatorname{Linear} (\boldsymbol{f}_\mathrm{M}), 
\end{equation}
where $H_{\text{head}}$ is the number of independent heads. 

Typically, these heads perform the same scale attention and thus lack respective diversity. To address this problem, we propose to split the heads evenly into two branches, each containing $\frac{H_{\text{head}}}{2}$ heads. In each branch, the input is first projected into the Query($Q$), Key($K$), and Values($V$). To capture hybrid-scale and multi-respective field interactions, we propose to utilize a hybrid-scale attention mechanism \cite{ren2023sg}, which further divides $\left\{Q, K, V\right\}$ into windows, producing a partition set $\left\{Q_i, K_i, V_i\right\}$ for $i=1,2,\cdots, n_{\text{win}}$. In contrast, the global attention block preserves the receptive field without partitioning. The feature of the local attention group is computed as follows:
\begin{eqnarray}
\operatorname{Atten}_i(Q_i, K_i, V_i)&=&\operatorname{softmax}\left(\frac{Q_i K_i^T}{\sqrt{d_k}}\right),\\
h_i &=& \operatorname{Atten}_i \cdot V_i,\\
h_{\text{local}} &=& \operatorname{Concat}(h_1, \cdots, h_i),
\end{eqnarray}
where $\operatorname{Atten}_i$ denote the attention score for each partition $i$, $\sqrt{d_k}$ represents the dimension of key, and $\operatorname{softmax}(\cdot)$ is the activate function. The hidden features of the global attention branch $h_{\text{global}}$ can be obtained following the above process without the need for window partitioning.  

Therefore, the final output $f_{\text{output}} \in \mathbb{R}^{1\times C}$can be obtained by concatenating the local and global features before being processed by an FFN layer as follows:
\begin{equation}
    f_{\text{output}} = \operatorname{Linear}(\operatorname{Concat}(h_{\text{local}}, h_{\text{global}})).
\end{equation}
This model enables the efficient fusion of diverse data, facilitating robust trajectory planning.

\begin{algorithm}[t]
	\caption{MSHA-RL Algorithm}\label{alg:alg1}
	\begin{algorithmic}[1]
		% \STATE 
		% \STATE {\textsc{TRAIN}}$(\mathbf{X} \mathbf{T})$
		\STATE \textbf{Input:} initial policy parameters $\theta_0$, initial policy parameters $\phi_0$, optimizer parameters, the maximum number of steps, and initial state $\bm{o}(0)$;
		\STATE \textbf{Output:} The MSHA-RL policy

		\FOR {episode = 0,2, ... , $K$}
		\FOR{step = 0,2,...$T$}
		\STATE The agent samples the action with $\pi_{\theta} \left({a}(t) | \bm{o}(t)\right)$
		\STATE The UAM executes the flying action.
		\STATE Compute current reward $r(t)$ and new state $\bm{o}(t+1)$.
		\ENDFOR
		\STATE Collect set of trajectory $\mathcal{I}_k = \{\iota_i\}_{i=0}^T$.
        \STATE Compute the cumulative discounted reward $R^\prime(t)$.
		\STATE Compute advantage estimates $\hat A(t)$ based on the current value function $V_\phi(\boldsymbol{o})$.
		\STATE Update the policy by maximizing the PPO clipped objective as given in Eq.~\eqref{eq:policy_loss}:
        $$\theta_{k+1} \gets \arg \max_{\theta}\left[ L_{\text{pf}}(\theta_{k})\right],$$
        by stochastic gradient ascent with Adam.
        \STATE Fit the value function by regression on mean-squared error loss as defined in Eq.~\eqref{eq:value_loss}:
        $$\phi_{k+1} \gets \arg \min_{\phi}\left[ L_{\text{vf}}(\phi_{k})\right],$$
        by stochastic gradient descent with Adam.
		\ENDFOR
		
	\end{algorithmic}
	\label{alg1}
\end{algorithm}
\subsection{Decision Module}
As shown in Fig.~\ref{MSHA}, the output of the MSHA is then fed into a Proximal Policy Optimization (PPO) module, which utilizes a policy network with parameters $\theta$ and a value network with parameters $\phi$ to inform decision-making. The policy network has a three-layer FFN and a $\operatorname{softmax}$ function as the output layer, which outputs the action probability distribution. The policy is updated by maximizing a clipped objective function:
\begin{equation}\label{eq:policy_loss}
L_{\text{pf}}=\mathbb{E}_t\left[\min \left(r_{\theta} \hat{A}(t), \operatorname{clip}\left(r_{\theta}, 1-\epsilon, 1+\epsilon\right) \hat{A}(t)\right)\right],
\end{equation}
where $r_{\theta} = \frac{\pi_{\theta}(a|\boldsymbol{o})}{\pi_{{\theta_{{\text{old}}}}}(a|\boldsymbol{o})}$ denotes the ratio of new to old policy probabilities while $\epsilon$ represents the range for clip.  Furthermore,  $\hat{A}(t)$ is the estimated advantage expressed as:
\begin{equation}
\hat{A}(t) = r(t) + \eta V_{\phi}\left(\boldsymbol{o}(t+1) \right) - V_{\phi} \left( \boldsymbol{o}(t)\right),
\end{equation}
where $V_{\phi}\left (\boldsymbol{o}\right)$ is the value predicted by the value network for state $\boldsymbol{o}$, and $\eta$ is a discount factor.

The value network shares the same structure as the policy network but without the output activation layer. It estimates expected future rewards and its loss function is defined as:
\begin{equation}\label{eq:value_loss}
    L_{\text{vf}}(\phi) = \mathbb{E}_t\left[\left(V_{\phi}\left(\boldsymbol{o}(t)\right)-R_{\text{cd}}(t)\right)^2\right],
\end{equation}
where $R_{\text{cd}}(t)$ represents the cumulative discounted reward at time step $t$, expressed as:
\begin{equation}
    R_{\text{cd}}(t) = \sum_{k=0}^{T-t}[\zeta^k \cdot r(t+k)],
\end{equation}
with $\zeta$ being a discount factor.

Based on the setup described above, the complete algorithm is presented in Algorithm \ref{alg:alg1}. The algorithm begins with initializing the policy parameters, comprising the policy network and value network, before entering the training phase. A mini-batch of features from multiple sources is extracted and processed by the MSHA model, with the output serving as the initial state during training. The UAM then samples actions based on the policy $\pi_{\theta} \left({a}(t) | \bm{o}(t)\right)$. Upon executing the action, we receive the current reward, leading to a transition to the subsequent state. These iterative operations continue until the time slot $T$ is reached (lines 4-8). When a set of trajectories is collected, we compute the cumulative discounted reward and the estimated advantage (lines 10-11). Based on these values, we update the policy network by maximizing the clipped objective and subsequently, refine the value function by minimizing its loss function (lines 12-13), forming the core mechanics of our integrated system's reinforcement learning process. Finally, we obtain the optimal MSHA-RL policy.

\begin{table}[t]
    \centering
    \caption{Summary of Variables in the Simulation}
    \begin{tabular}{@{}lll@{}}
        \toprule
        \textbf{Category} & \textbf{Variables} & \textbf{Value} \\
        \midrule
        \multirow{4}{*}{UAM} & Flight altitude & $100~\mathrm{m}$ \\
        & Flight speed & $120~\mathrm{km/h}$ \\
        &\makecell[l]{Available seats} & $2$ \\
        \midrule
        \multirow{4}{*}{Communication} & Hight of GBS & $15~\mathrm{m}$ \\
        & Carrier frequency & $2~\mathrm{GHz}$ \\
        & \makecell[l]{Isotropic antenna gain}& $0~\mathrm{dB}$ \\
        \midrule
        \multirow{3}{*}{Scenario 1} & Berlin map size & $3.2 \times 3.2~\mathrm{km}$  \\
        & \makecell[l]{Number of passengers} & $4$  \\
        % & The distribution of arrival time& $\mu=0, \sigma=0.01$\\
        \midrule
        \multirow{3}{*}{Scenario 2}& Detroit map size & $11 \times 11~\mathrm{km}$  \\
        & \makecell[l]{Number of passengers} & $5$  \\
        \midrule
         \multirow{3}{*}{\makecell[l]{Algorithm\\Hyperparameters}} &Learning rate& $10^{-5}$ \\
        & Total steps & $2500$ \\
        & Batch size& $400$ \\
        \bottomrule
    \end{tabular}
\end{table}

\section{Experiment Results and Analyses}\label{sec:experiment}
\subsection{Experiment Settings}
\begin{figure*}[htbp]
	\begin{center}
		\includegraphics[width=\textwidth]{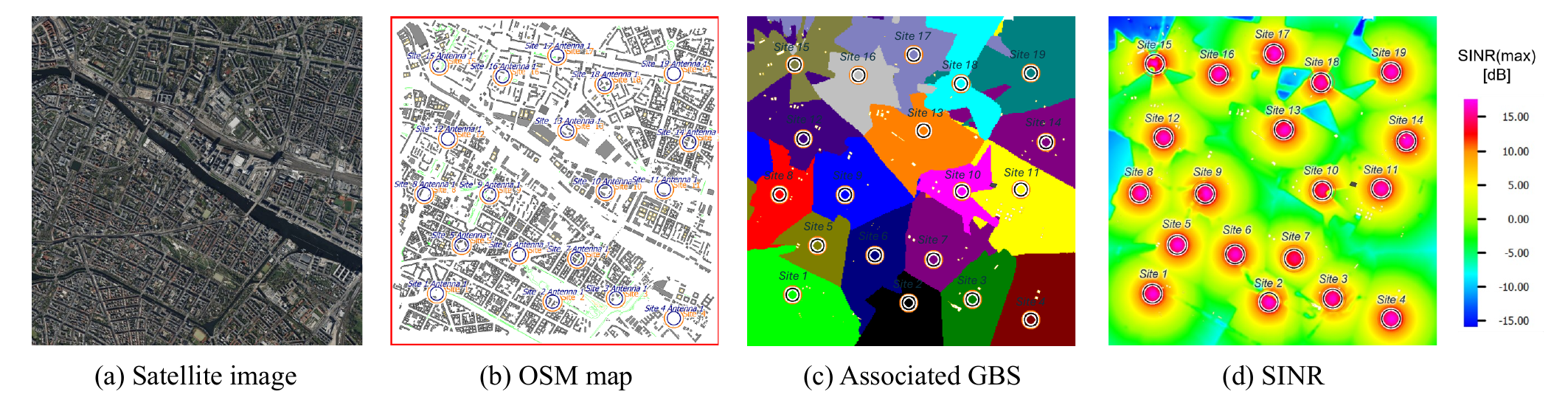}
	\end{center}
        \vskip -1.3em
	\caption{Radio Map Construction, Berlin}
	\label{berlin_sinr}
    \vskip -1em
\end{figure*}

\begin{figure*}[htbp]
	\begin{center}
		\includegraphics[width=\textwidth]{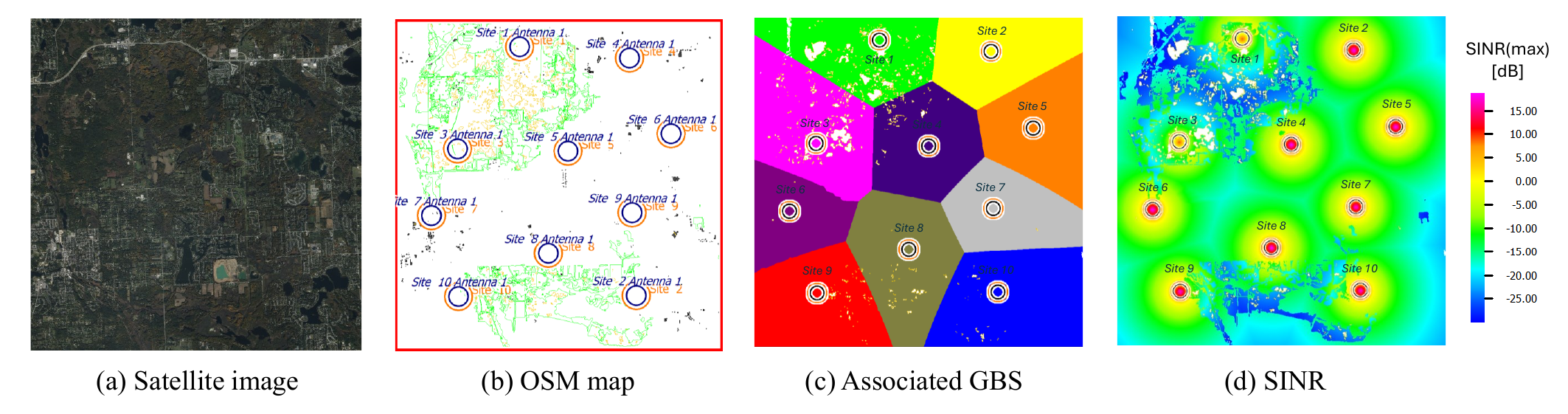}
	\end{center}
        \vskip -1.3em
	\caption{Radio Map Construction, Detroit}
	\label{detroit_sinr}
\vskip -2em
\end{figure*}

To simulate a real environment, we extracted two maps from the Open Street Map (OSM), incorporating height information. The first map depicts an urban area of Berlin, Germany, with dimensions of $3.2~\mathrm{km} \times 3.2~\mathrm{km}$, while the second shows a suburban area of Detroit, USA, covering $11~\mathrm{km} \times 11~\mathrm{km}$. We then constructed radio maps for both areas. The radio maps for these two scenarios are shown in Figs.~\ref{berlin_sinr} and ~\ref{detroit_sinr}, respectively. The satellite images presented in Figs.\ref{berlin_sinr}(a) and ~\ref{detroit_sinr}(a) show that the selected Berlin scenario has many residential buildings, while the suburban area of Detroit is covered with a larger area of forest. Therefore, the number and distribution of GBSs differ significantly between the Berlin and Detroit areas, as illustrated in Fig.~\ref{berlin_sinr}(b) and Fig.~\ref{detroit_sinr}(b), respectively. In both areas, the GBSs are positioned at a height of $15\mathrm{m}$. 

We next imported maps into WinProp software for radio propagation simulation. The simulation environment is configured by setting the geographical location of the base station and defining relevant parameters such as antenna height, transmission power, and frequency. With these settings in place, we generated the simulated radio maps of two scenarios, which are illustrated in Figs.~\ref{berlin_sinr}(c),~\ref{berlin_sinr}(d) and Figs.~\ref{detroit_sinr}(c),~\ref{detroit_sinr}(d), respectively.

We assume that the UAM is equipped with isotropic antennas with a gain of unity. The large-scale channel gain between GBS and the UAM is modeled based on the 3GPP technical report for the Urban Micro (UMi) scenario\cite{3gpp2018lte}. Specifically, we employed the LOS and NLOS channel models at location $\boldsymbol{u}$, which are modeled as $\bar{h}_m^{\mathrm{LoS}}$ and $\bar{h}_m^{\mathrm{NLoS}}$ \cite{chen2024optimal}:  
\begin{equation}
\begin{aligned}
\bar{h}_m^{\mathrm{LoS}}(\boldsymbol{u})= & \frac{G_m(\boldsymbol{u})}{2}+\frac{1}{2} \min \left\{\bar{h}_m^{\mathrm{FS}},-30.9-(22.25\right. \\
& \left.\left.-0.5 \log _{10} \bar{H}\right) \log _{10} d_m(\boldsymbol{u})-20 \log _{10} f_c\right\},
\end{aligned}
\end{equation}

\begin{equation}
\begin{aligned}
\bar{h}_m^{\mathrm{NLoS}}(\boldsymbol{u})= & \frac{G_m(\boldsymbol{u})}{2}+\frac{1}{2} \min \left\{\bar{h}_m^{\mathrm{LoS}}(\boldsymbol{u}),-32.4-(43.2\right. \\
& \left.\left.-7.6 \log _{10} \bar{H}\right) \log _{10} d_m(\boldsymbol{u})-20 \log _{10} f_c\right\},
\end{aligned}
\end{equation}
where the isotropic gain of GBS is $G_m(\boldsymbol{u})=0~\mathrm{dB}$, the distance between the UAM and its serving GBS is denoted as $d_m$. Furthermore, the UAM flies at a fixed height of $\bar{H}=100~\mathrm{m}$ whereas the carrier frequency is $f_c=2~\mathrm{GHz}$. Finally, $\bar{h}_m^{\mathrm{FS}}$ denotes the free-space path loss.

According to Section \ref{sec:map}, we constructed the SINR maps of the selected area.
% 设置4个乘客，每个乘客动态出现，我们设乘客起点为S_i,D_i
Furthermore, we consider that the UAM flies at a constant speed of $120\mathrm{km/h}$ with two available seats and operates within the service area. 

Two distinct scenarios are implemented to evaluate the proposed algorithm under different spatial and temporal conditions: Berlin and Detroit. The Berlin scenario comprises four passengers at most, and each passenger is characterized by a unique origin-destination pair $(S_i, D_i)$. Specifically, the first three passengers ($i = 1,2,3$) arrive at the initial time, while the fourth passenger randomly arrives in different locations and time slots. Each time slot is defined as three seconds. The Detroit scenario extends both the service area and passenger complexity to five individuals. The first two passengers ($i = 1,2$) are characterized by early arrivals, while the subsequent passengers demonstrate progressively random arrival patterns.

\subsection{Experiment Evaluation}

To evaluate the performance of various methods, we define the following evaluation metrics:
\begin{itemize}[leftmargin=*]
    \item Total Distance (\textbf{TD}): The cumulative distance traveled by the UAM after completing all passenger pickups.
    \item Average Total Time (\textbf{ATT}): The average total time consumption per passenger, involving waiting time and time spent onboard the UAM.
    \item Average Waiting Time (\textbf{AWT}): The average duration a passenger waits from arrival at the airport until boarding the UAM.
    \item Empty-Loaded Rate (\textbf{ELR}): The percentage of time the UAM operates without passengers.
    \item Pick-Up Rate (\textbf{PR}): The percentage of successfully passenger pick-ups.
\end{itemize}
\begin{table*}[t]
% \vskip -1.3em
\caption{Metrics Comparison}
\label{tab:comparison}
\centering
\begin{tabular}{c | ccccc |ccccc}  % 使用 @{} 去掉列间的空格
\toprule
\multirow{2}{*}{Methods} &\multicolumn{5}{c|}{\textbf{Berlin}} &\multicolumn{5}{c}{\textbf{Detroit}} \\ 
\specialrule{0.0em}{0.0ex}{-0.1ex} % 减小下行间距
\cmidrule(lr){2-6} \cmidrule(lr){7-11} 
\specialrule{0.0em}{-0.5ex}{0.0ex} % 减小上行间距
&$\bar{\gamma}_{\mathrm{T}}(\mathrm{dB})$& TD($\mathrm{m}$)  &ATT($\mathrm{s}$)& AWT($\mathrm{s}$) & ELR(\%) 
                        &$\bar{\gamma}_{\mathrm{T}}(\mathrm{dB})$& TD($\mathrm{m}$)  &ATT($\mathrm{s}$)& AWT($\mathrm{s}$) & ELR(\%)\\ \midrule
\multirow{2}{*}{CPTSP}& -7   & 9025          &142.86 & 107.76&48.15 & -21&30006&502.75&379.46&32.22\\  
                      & -6   & 9207          &146.07 & 109.89&47.41 & -20&30006&502.75&379.46&32.22\\ 
                      & -5   &  9207         &146.07 & 109.89&47.41 & -19&30195&503.92&379.46&32.02\\ 
                      & -4   &  9225         &147.39 & 111.21&47.41 & -18&30272&510.8 &387.52&32.81\\ 
                      & -3   &  9425         &151.08 & 114.45&48.22 & -17&30458&511.97&387.52&32.61\\ \midrule
\multirow{2}{*}{PDPCC}& -7   &  7932         &131.25 & 90.12 &42.82 & -21&29443&502.36&360.55&26.69\\  
                      & -6   &  8073         &134.46 & 92.25 &42.07 & -20&29443&502.36&360.55&26.69\\  
                      & -5   &  8073         &134.46 & 92.28 &42.07 & -19&29632&503.53&360.55&26.52\\  
                      & -4   &  8132         &135.78 & 93.6  &42.07 & -18&29709&510.41&368.6&27.34\\ 
                      & -3   &  8332         &139.47 & 96.84 &43.16 & -17&29894&511.58&368.6&27.19\\ \midrule
MSHA-RL               & -7   & \textbf{5100} &\textbf{100.5}  & 56.25 & 40.39& -21&\textbf{24960}&\textbf{422.92}&248.32&31.64\\  
(OURS)                & -6   & \textbf{5600} &\textbf{112.5}  & 61.5  & 41.37& -20&\textbf{26560}&\textbf{465.6}&318.16&14.29\\
                      & -5   & \textbf{5700} &\textbf{113.25} & 63.75 & 37.93& -19&\textbf{26880}&\textbf{467.54}&291&28.24\\
                      & -4   & \textbf{5900} &\textbf{114.75} & 64.5 & 36.67 & -18&\textbf{27520}&\textbf{481.12}&302.64&27.59\\
                      & -3   & \textbf{6200} &\textbf{121.5}  & 69.75 & 36.6 & -17&\textbf{28160}&\textbf{498.58}&316.22&26.97\\
\bottomrule          
\end{tabular}
% \vskip -1.3em
\end{table*}

\begin{figure}[!htp]
\vskip -2em
\centering
      \subfloat[]{\includegraphics[width=0.4\textwidth]{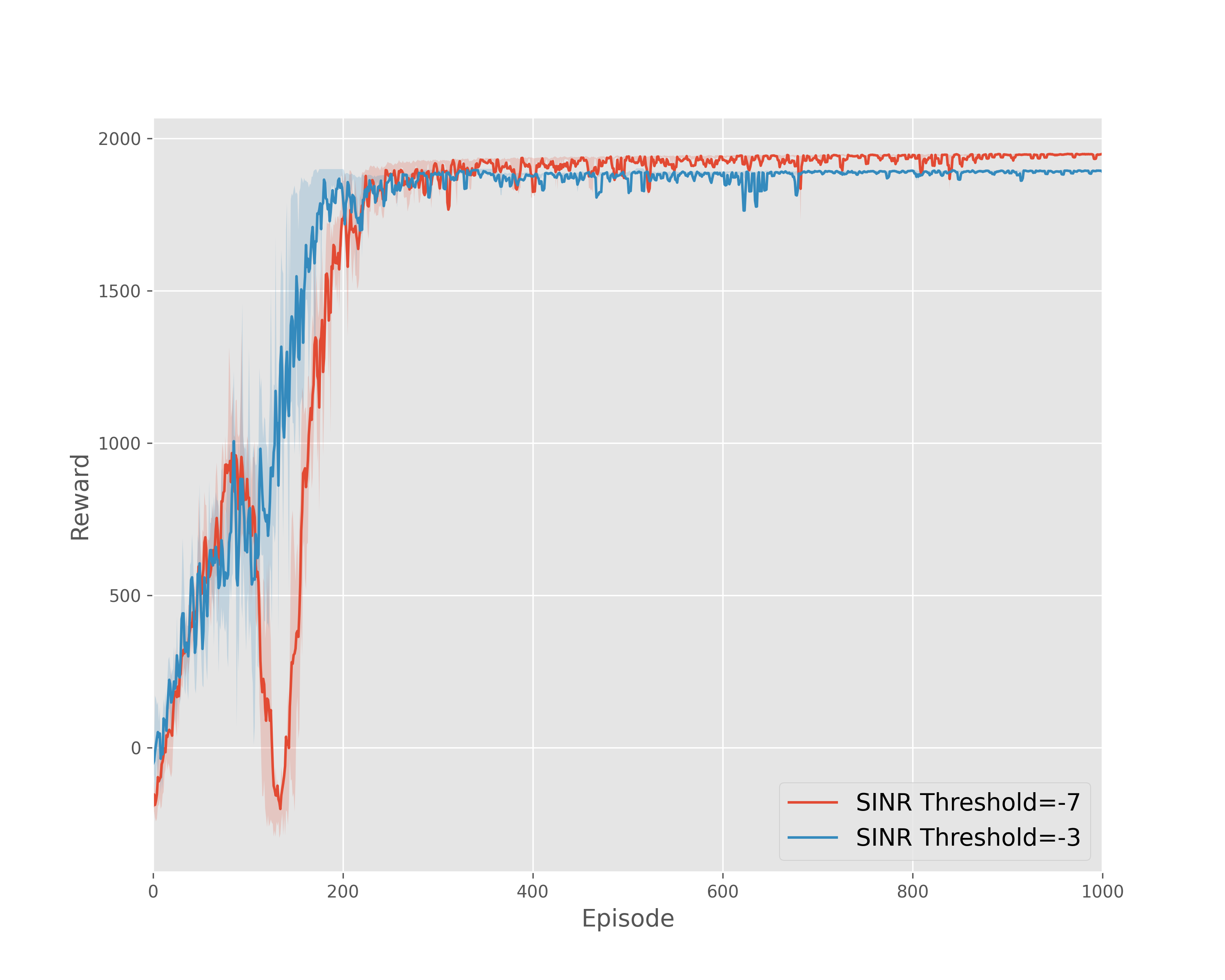}}
     \hfill 	
     % \vskip -0.5em
      \subfloat[]{\includegraphics[width=0.4\textwidth]{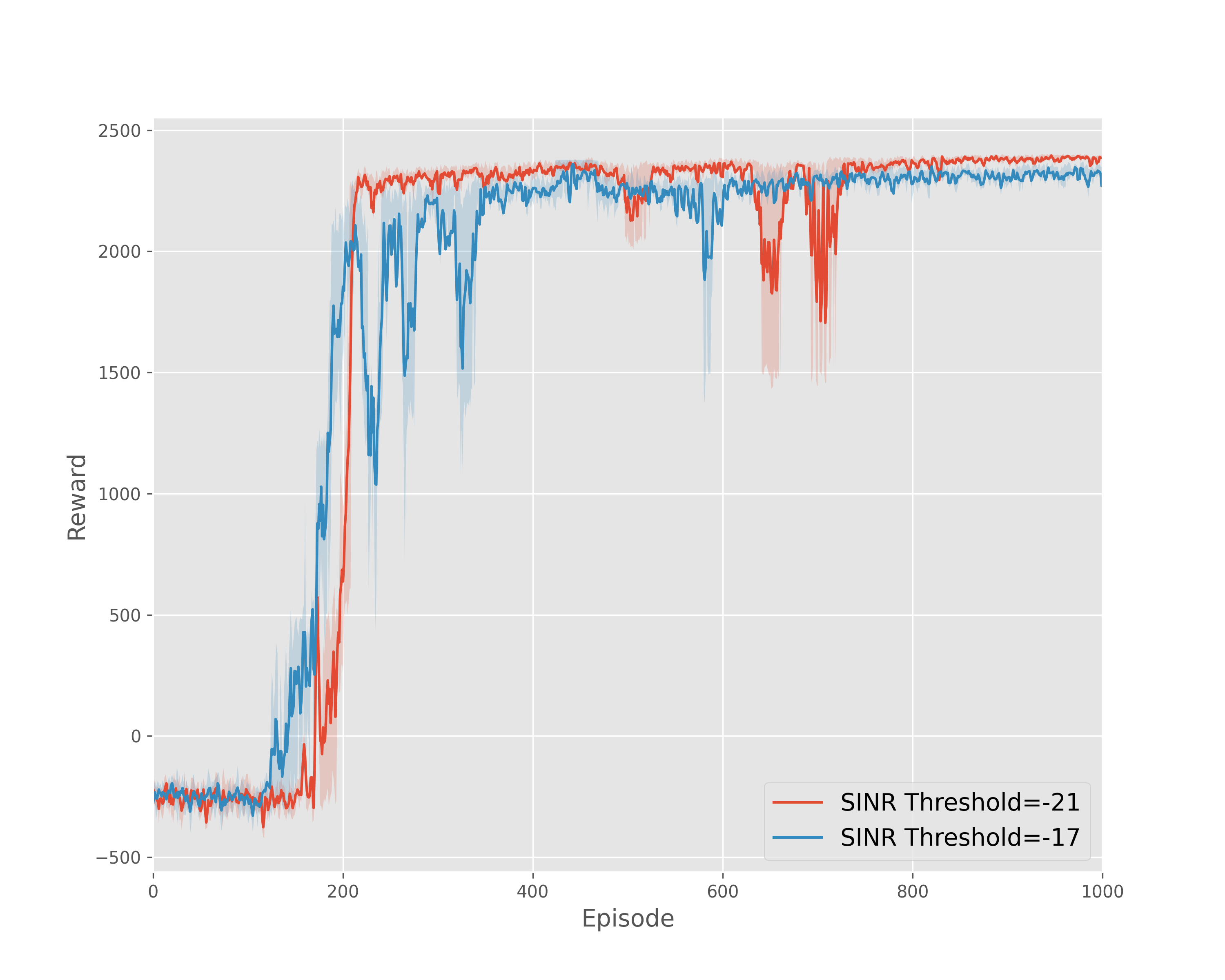}}
     \caption{Convergence performance achieved by MSHA-RL at various SINR thresholds.}
     \label{convergence}
\end{figure}
The experiments were performed in the Ubuntu 22.04 LTS operating system with an Intel 13900K CPU and an NVIDIA RTX3090 GPU.

\subsubsection{\textbf{Convergence Results}}
We conducted a series of experiments under two distinct scenarios, with SINR threshold values ranging from $-7$ to $-3~\mathrm{dB}$ for the Berlin scenario and $-21$ to $-17~\mathrm{dB}$ for the Detroit scenario.  These policies of both scenarios converged after $800$ episodes, as shown in Fig.~\ref{convergence}. Additionally, their inference times consistently remained below $1$ second. 

\subsubsection{\textbf{Metrics Comparison}}
In the following experiments, we will benchmark our proposed MSHA-RL against two conventional approaches, namely CPTCP and PDPCC, proposed in \cite{chen2024optimal} to solve the trajectory planning problem. Both methods involve a two-step process: determining the optimal task completion sequence and generating a specific path using the Dijkstra algorithm. The CPTSP requires completing the current passenger's pickup and drop-off before proceeding to the next passenger, while PDPCC allows for ride-sharing routes. In Table.~\ref{tab:comparison}, we present comprehensive quantitative metrics of three distinct methods, evaluating across two scenarios with distinct SINR threshold ranges. 

First of all, since all methods achieve a precision-recall (PR) of $100\%$, this metric has been omitted from Table~\ref{tab:comparison} for clarity and conciseness. It is evident from Table.~\ref{tab:comparison} that the TD of the path generated by the proposed MSHA-RL is significantly shorter than those produced by the two conventional algorithms for both scenarios, regardless of the communication limits. This is because our proposed algorithm quickly responded to passengers' ride requests, allowing dynamic adjustments to the pickup sequence and obtaining a more optimized path. Meanwhile, as the minimum SINR requirement increased, TD gradually increased to avoid areas with poor signal quality. Furthermore, since the speed is constant, the ATT of the MSHA-RL is minimized compared to CPTSP and PDPCC. Although the flight time of some users was increased due to the carpooling strategy, the overall efficiency of the system has been greatly improved. This optimal scheduling and path planning also led to the lowest AWT and enhances the user experience in general.

Finally, ELR indicates the proportion of time that the UAM remains unoccupied. Thus, a lower ELR corresponds to higher UAM utilization. As shown in Table.~\ref{tab:comparison}, the proposed MSHA-RL achieved the highest UAM utilization across all SINR threshold ranges. The implemented policy demonstrates robust performance optimization even with increased service coverage area and passenger demand complexity, successfully minimizing both path lengths and overall travel time. 

\subsubsection{\textbf{Visualized Results}}
\begin{figure*}[htbp]
	\begin{center}
		\includegraphics[width=\textwidth]{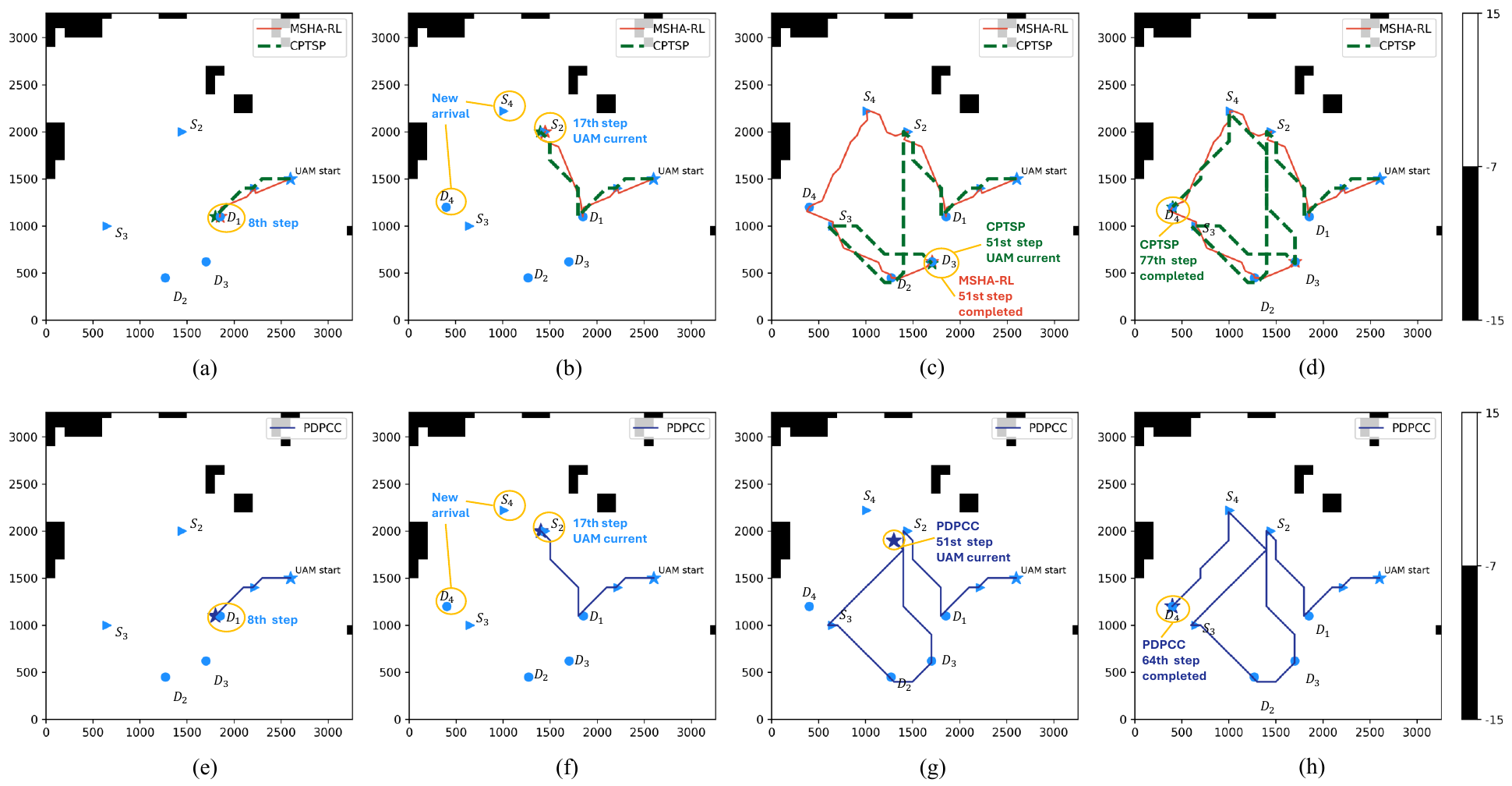}
	\end{center}
% \vskip -1em
	% \caption{Real-time trajectory visualizations at various time steps: (a) 8th time step; (b) 17th time step; (c) 51st time step; (d) 77th time step.}
    \caption{Real-time trajectory visualizations at various time steps.}
	\label{step_figure}
% \vskip -1em
\end{figure*}

\begin{figure*}[htp] 
    \centering
    \subfloat[MSHA-RL and Straight Line\label{sinr_-3_a}]{
        \includegraphics[width=0.32\textwidth]{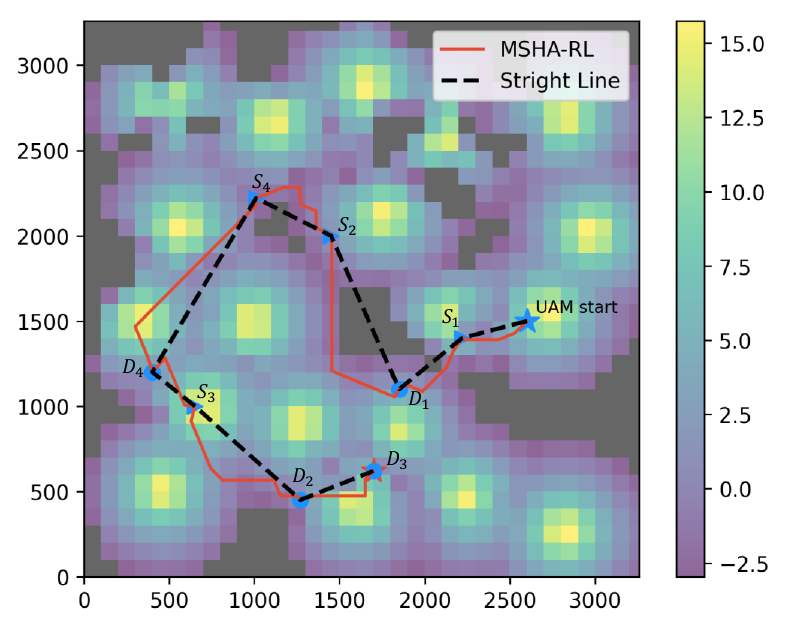}
    }
    \hfill
    \subfloat[CPTSP\label{sinr_-3_b}]{
        \includegraphics[width=0.32\textwidth]{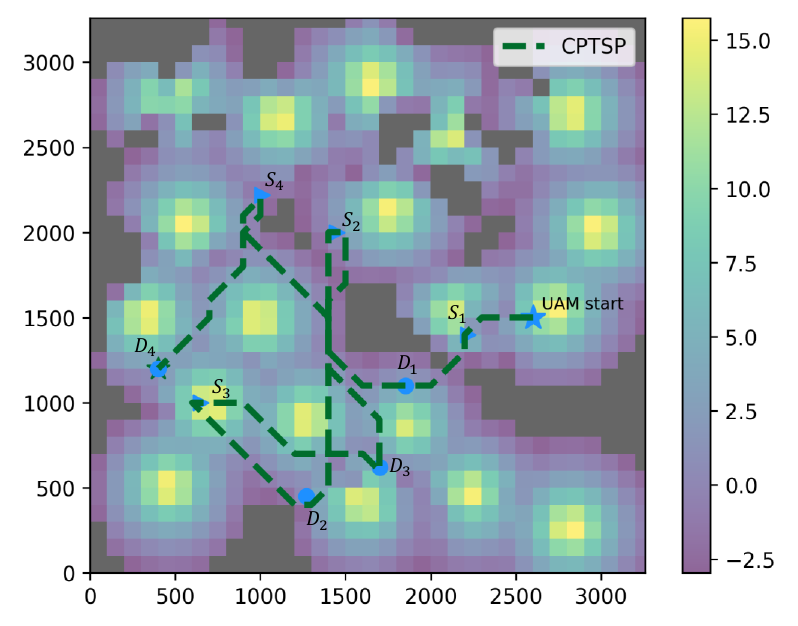}
    }
    \hfill
    \subfloat[PDPCC\label{sinr_-3_c}]{
	\includegraphics[width=0.31\textwidth]{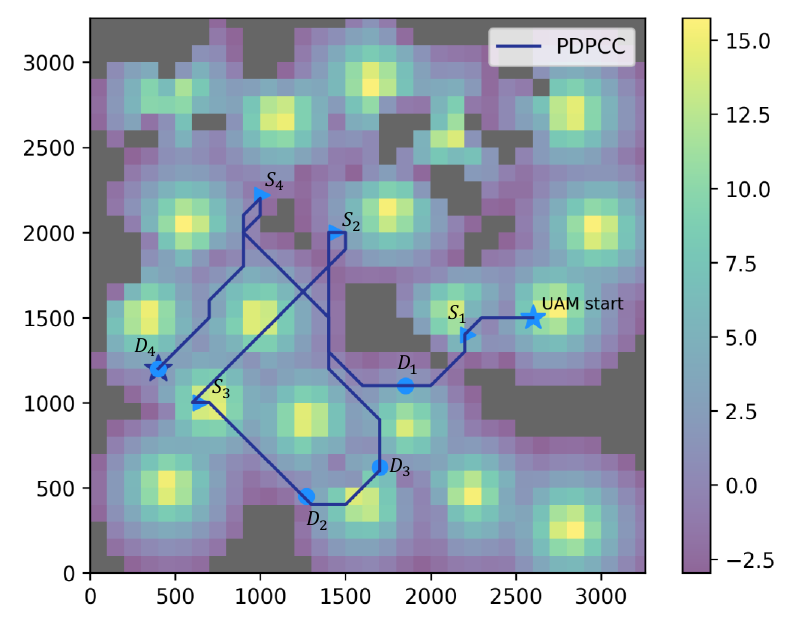}
    }
    \caption{The UAM Trajectory generated by different methods at $\bar{\gamma}_{\mathrm{T}}=-3~\mathrm{dB}$, Berlin.}
    \label{sinr_-3}
\end{figure*}

\begin{figure*}[!htp]
    \centering
      \subfloat[MSHA-RL and Straight Line]{\includegraphics[width=0.33\textwidth]{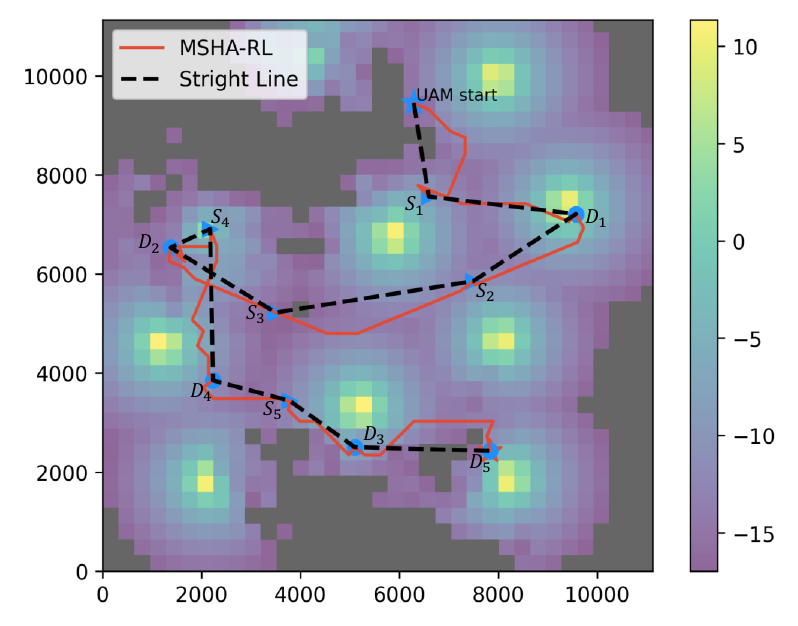}}
     \hfill 	
      \subfloat[CPTSP]{\includegraphics[width=0.33\textwidth]{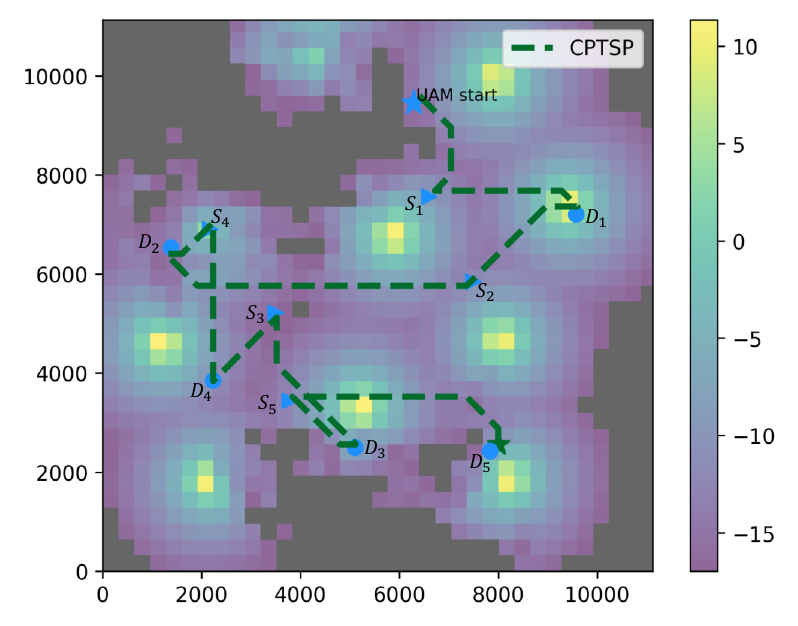}}
       \hfill 	
      \subfloat[PDPCC]{\includegraphics[width=0.33\textwidth]{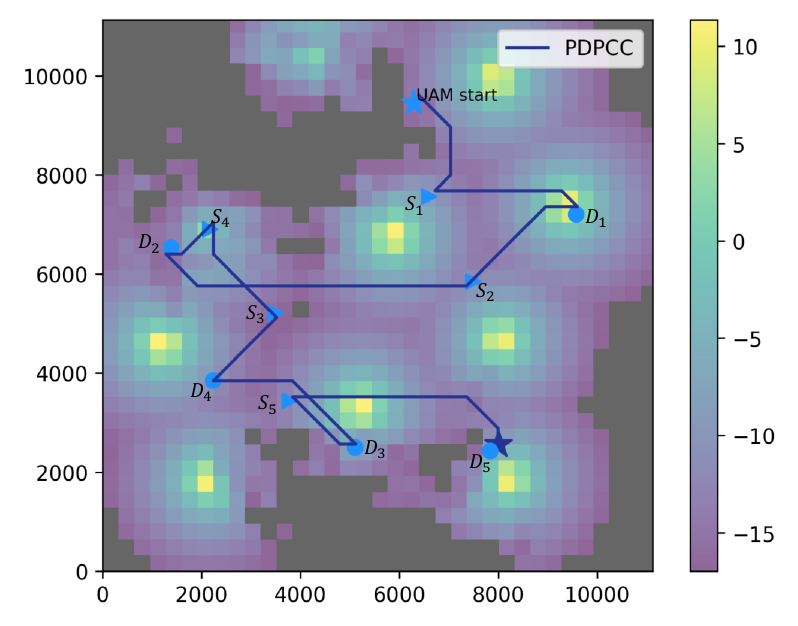}}
    % \vskip 1em
    \caption{The UAM Trajectory generated by different methods at $\bar{\gamma}_{\mathrm{T}}=-17~\mathrm{dB}$, Detroit.}
    \label{sinr_-17}
\end{figure*}

To intuitively illustrate our approach, Fig.~\ref{step_figure} presents an example of the real-time UAM trajectory in the Berlin scenario. The comparative visualization depicts the specific path of multiple approaches across distinct time steps, with an SINR threshold of $\bar{\gamma}_{\mathrm{T}} = -7~\mathrm{dB}$. At different times, there are different passengers arriving. From Figs.~\ref{step_figure}(a) and (b), Passengers 1, 2, and 3 submitted the ride requests at the very beginning, and Passenger 4 did not send its ride request until the $17$-th time step of the flight. The proposed MSHA-RL (marked in red) adjusts the path in real-time to prioritize the pickup of Passenger 4, while the offline methods adhere to the pre-determined task sequence, serving Passenger 3, as shown in Fig.~\ref{step_figure}(c). Specifically, the performance of each method is as follows:
\begin{itemize}[leftmargin=*]
\item MSHA-RL: the UAM completed the shuttle task for all passengers by $T=51$, following the MSHA-RL path planning strategy. The task sequence was as follows: 
start-$S_1$-$D_1$-$S_2$-$S_4$-$D_4$-$S_3$-$D_2$-$D_3$;
\item PDPCC: In Fig.~\ref{step_figure}(d), the UAM successfully completed the task at $T=64$, employing the PDPCC (marked in blue) strategy, which generated the following task sequence: start-$S_1$-$D_1$-$S_2$-$S_3$-$D_2$-$D_3$-$S_4$-$D_4$;
\item CPTSP: In contrast, the CPTSP (marked in green) strategy required $13$ {\em additional} steps to complete all tasks, finally finishing at $T=77$, resulting in the following task sequence: start-$S_1$-$D_1$-$S_2$-$D_2$-$S_3$-$D_3$-$S_4$-$D_4$.
\end{itemize}

The above results reveal that the conventional algorithms can only perform the pick-up of Passenger 4 after completing the initial pick-up of Passengers 1, 2, and 3 first. As a result, these conventional algorithms failed to respond to Passenger 4 in real time, resulting in a longer trip and longer consumption time. Meanwhile, in contrast to CPTSP, which can only pick up one passenger at a time, PDPCC achieves better path planning by considering carpooling routes.

\begin{table*}[ht]
	\vskip 1em
	\centering
	\caption{ABLATIVE EVALUATIONS FOR METHODS}
	\begin{tabular}{c | ccccc | cccccc}
		\toprule
		Scenario                    &Model         & LE        & DE       & MSFea  &HAF       &PR(\%) & TD($\mathrm{m}$)  
		&ATT($\mathrm{s}$)& AWT($\mathrm{s}$) & ELR(\%) & Task Status \\ \midrule
		\multirow{4}{*}{Berlin}     &1 & \ding{52} & \ding{56} & \ding{56} & \ding{56}  &25& -    & -     & -     & - &Failure        \\
		&2 & \ding{56} & \ding{52} & \ding{56} & \ding{56}  &75& -    & -    & -    & - &Failure      \\
		&3 & \ding{56} & \ding{52} & \ding{52} & \ding{56}  &100&-    &-     &-     &-    &Failure      \\
		&4 & \ding{56} & \ding{52} & \ding{52} & \ding{52}  &100&6200     &121.5     &69.75     &36.6  &Success\\ \midrule  
		\multirow{4}{*}{Detroit}    &1 & \ding{52} & \ding{56} & \ding{56} & \ding{56}  &20& -    & -    & -    & -  &Failure       \\
		&2 & \ding{56} & \ding{52} & \ding{56} & \ding{56}  &80& -    & -    & -    & -   &Failure      \\
		&3 & \ding{56} & \ding{52} & \ding{52} & \ding{56}  &100&36160     &0.4474     &706.16    &403.52 &Success        \\
		&4 & \ding{56} & \ding{52} & \ding{52} & \ding{52}  &100&28160     &552.9     &370.54     &26.97 &Success \\
		\bottomrule
	\end{tabular}
	\label{table:ablation}
\end{table*}

\begin{figure}[thp]
	\vskip -1em
	\centering
	\subfloat[]{\includegraphics[width=0.4\textwidth]{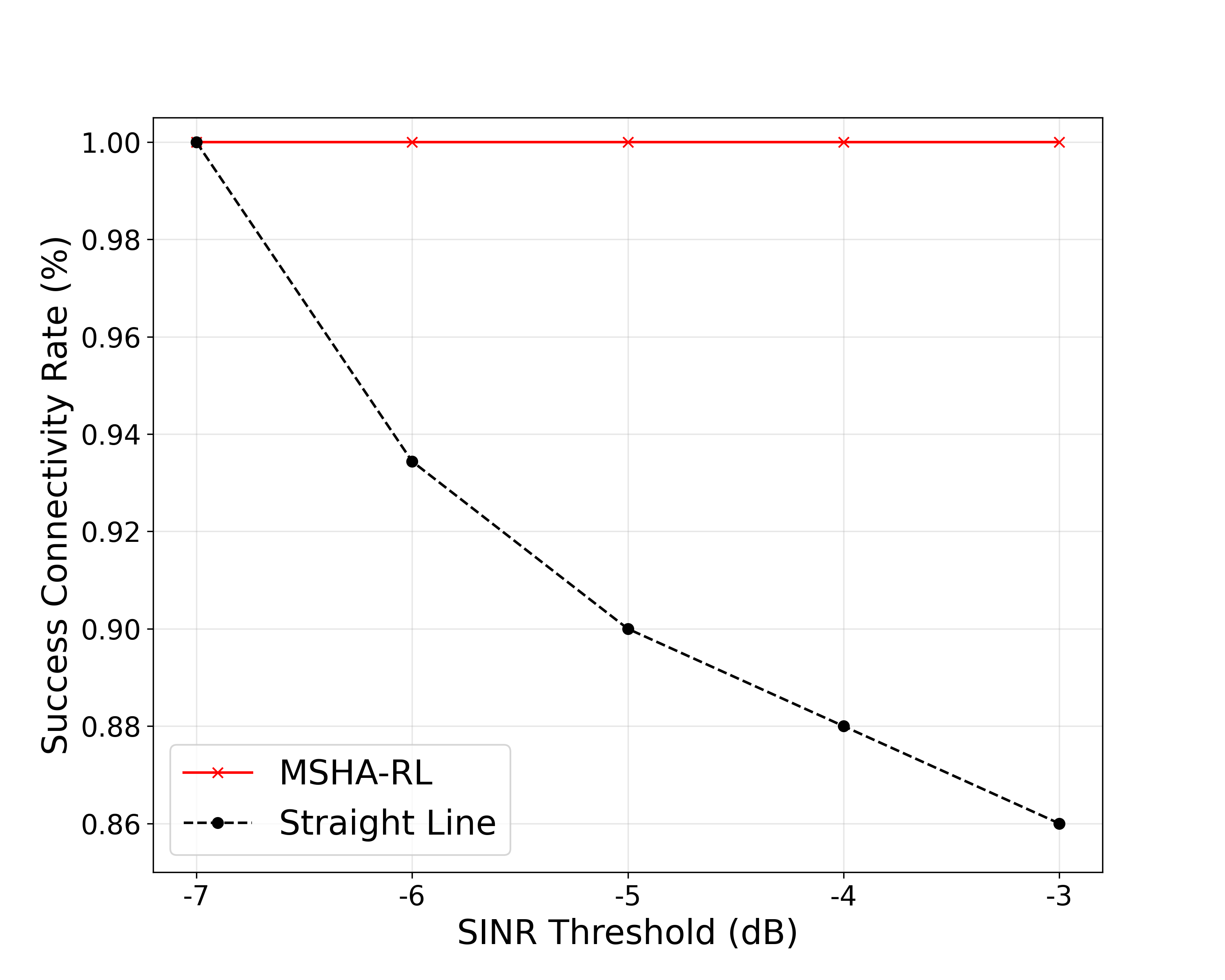}}
	\hfill 	
	\vskip -0.2em
	\subfloat[]{\includegraphics[width=0.4\textwidth]{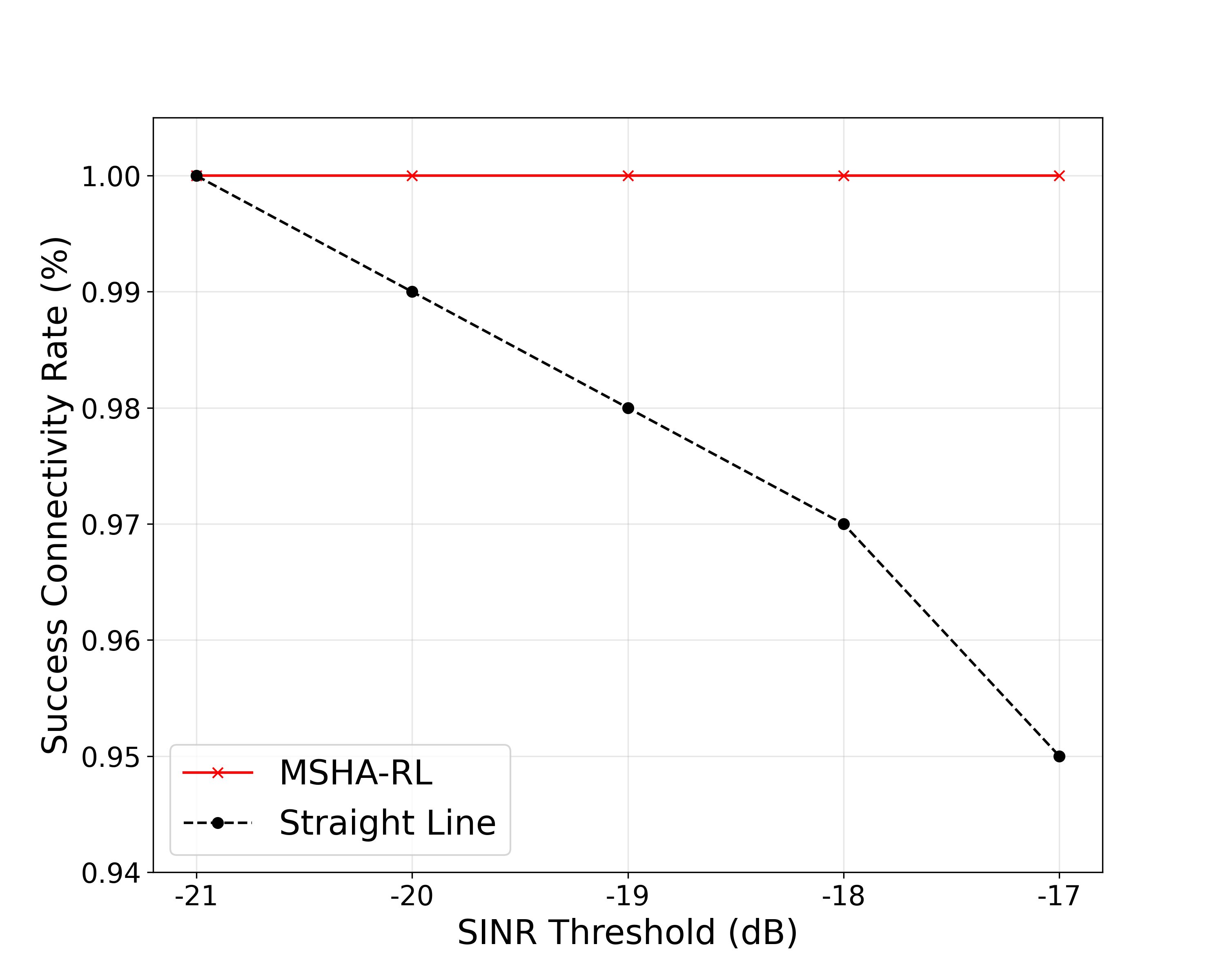}}
	\caption{The communication connectivity}
	\label{com_connectivity}
    \vskip -1em
\end{figure}
\subsubsection{\textbf{Connectivity Analysis}}
Next, we visualize the trajectory planning results of various methods under stringent communication conditions, with an SINR threshold of $\bar{\gamma}_{\mathrm{T}} = -3~\mathrm{dB}$ for the Berlin scenario, and $\bar{\gamma}_{\mathrm{T}} = -17~\mathrm{dB}$ for the Detroit scenario, as depicted in Figs.~\ref {sinr_-3} and \ref {sinr_-17} respectively. The results obtained using our proposed method and the straight-line method are shown in Figs.~\ref{sinr_-3}(a) and \ref{sinr_-17}(a), while Figs.~\ref{sinr_-3}(b) and \ref{sinr_-17}(a) display the outcomes of two conventional approaches for comparison. The task sequence of UAM under the Berlin scenario is provided above, and that of the Detroit scenario is demonstrated as follows:
\begin{itemize}[leftmargin=*]
\item MSHA-RL: start-$S_1$-$D_1$-$S_2$-$S_3$-$D_2$-$S_4$-$D_4$-$S_5$-$D_3$-$D_5$;
\item PDPCC:  start-$S_1$-$D_1$-$S_2$-$D_2$-$S_4$-$D_4$-$S_3$-$D_3$-$S_5$-$D_5$;
\item CPTSP:  start-$S_1$-$D_1$-$S_2$-$D_2$-$S_4$-$S_3$-$D_4$-$D_3$-$S_5$-$D_5$.
\end{itemize}

It is evident that the proposed MSHA-RL has significantly reduced the travel distance required to complete all tasks compared to traditional methods, while avoiding the region with SINR values below the threshold compared with the straight-line method. MSHA-RL dynamically adjusted its strategy to accommodate changing user needs and, by considering multiple action directions, generated a smoother and more efficient path.

Fig.~\ref{com_connectivity} compares the communication connectivity of UAMs following straight-line flight paths with those using MSHA-RL under varying SINR constraints. Figs.~\ref{com_connectivity}(a) and~\ref{com_connectivity}(b) illustrate the results for the Berlin and Detroit scenarios, respectively. The straight-line flight path followed the same optimal task sequence as the proposed MSHA-RL. The results demonstrate communication connectivity under five distinct SINR thresholds. As the constraints become more stringent, the flight paths generated by MSHA-RL maintain $100\%$ connectivity across all constraints in both simulation scenarios. In contrast, the communication connectivity of UAM flying in a straight line exhibited degraded connectivity performance, reaching the minimum of $86\%$ and $95\%$ in respective scenarios, which failed to meet the communication reliability for robust UAM operations. 

\subsection{Ablation Study}
To evaluate the impact of individual components on the performance of the proposed MSHA-RL policy, we conducted a comprehensive ablation study across two scenarios: Berlin with a SINR threshold of $\bar{\gamma}_{\mathrm{T}} = -3~\mathrm{dB}$, and Detroit with $\bar{\gamma}_{\mathrm{T}} = -17~\mathrm{dB}$. The PR represents the pick-up rate of passengers. If PR is below 100\%, the task is considered a failure, rendering other indicators meaningless; otherwise, all indicator results will be reported.

As shown in Table~\ref{table:ablation}, Model 1, which employs only a Linear Encoder (LE) for information processing, demonstrated significantly compromised performance, with only $25\%$ and $20\%$ PR in the respective scenarios. The introduction of the Designed Encoder (DE) in Model 2 exhibited increased passenger pick-up rates compared to Model 1, yet failed to achieve mission completion. While the integration of the Multi-Source Feature (MSFea) module manifested substantial improvements through enhancing multi-source information alignment, the model's performance lacked robustness. Specifically, in the Berlin scenario, the model failed to complete the mission. Furthermore, the generated trajectories in the Detroit scenario exhibited suboptimal performance across all evaluation metrics—TD, ATT, AWT, and ELR.

The final architecture, incorporating DE, MSFea, and Hybrid Attention Fusion (HAF) components, achieved superior performance across all quantitative metrics. This comprehensive model effectively fuses multi-source features while capturing the global and local attention of the features to express the observation better. This ablation study underscores the critical role of the MSHA model's modules, particularly the MSFea and HAF modules, in achieving optimal system performance in both Berlin and Detroit complex environments.

\section{Conclusion}\label{sec:conclude}
This work has studied the real-time ride-sharing trajectory planning for UAM under communication connectivity constraints. To minimize overall travel time costs while ensuring reliable communication, we have developed a comprehensive system model by constructing radio maps and formulating the MDP problem for ride-sharing. Our proposed MSHA-RL architecture, incorporating a multi-source fusion and a hybrid attention module, can effectively align multi-source information with a large dimension gap and obtain the feature output from global and local perspectives. 
Extensive experimental results have shown the proposed MSHA-RL can significantly outperform conventional methods. In the Berlin scenario, the proposed MSHA-RL achieved average reductions of up to $3,501$ meters in UAM travel distance TD and $34.18$ seconds in average total time consumption of the passenger, ATT. The performance advantages were even more pronounced in the more complex Detroit scenario, where the MSHA-RL decreased the TD by $7,144$ meters and ATT by $40$ seconds. 

% These improvements are attributed to MSHA-RL’s capability for real-time path planning, effectively minimizing travel distance while maintaining reliable communication.

% \small
\bibliographystyle{IEEEtran}
\bibliography{refs}

\end{document}